\RequirePackage{fix-cm}
\documentclass[twocolumn]{svjour3}          
\smartqed 
\usepackage{graphicx}
\usepackage{hyperref}
\usepackage{marvosym}
\usepackage{todonotes}

\journalname{Int J Soc Robot}

\begin{document}

\title{Knowledge-Grounded Dialogue Flow Management for Social Robots and Conversational Agents
}



\author{Lucrezia Grassi         \and
        Carmine Tommaso Recchiuto \and 
        Antonio Sgorbissa
}


\institute{L. Grassi (\Letter) \and C.T. Recchiuto \and A. Sgoribissa \at
              University of Genoa, DIBRIS, via all'Opera Pia 13 \\
              Tel.: +39 010 3532801\\
              Fax:  +39 010 3532154\\
             \email{\href{mailto:lucrezia.grassi@edu.unige.it}{lucrezia.grassi@edu.unige.it}}  \\             
}

\date{Received: date / Accepted: date}

\maketitle

\begin{abstract}
The article proposes a system for knowledge-based conversation designed for Social Robots and other conversational agents. The proposed system relies on an Ontology for the description of all concepts that may be relevant conversation topics, as well as their mutual relationships. The article focuses on the algorithm for Dialogue Management that selects the most appropriate conversation topic depending on the user's input. Moreover, it discusses strategies to ensure a conversation flow that captures, as more coherently as possible, the user's intention to drive the conversation in specific directions while avoiding purely reactive responses to what the user says. 
To measure the quality of the conversation, the article reports the tests performed with 100 recruited participants, comparing five conversational agents: (i) an agent addressing dialogue flow management based only on the detection of keywords in the speech, (ii) an agent  based both on the detection of keywords and the Content Classification feature of Google Cloud Natural Language, (iii) an agent that picks conversation topics randomly, (iv) a human pretending to be a chatbot, and (v) one of the most famous chatbots worldwide: Replika. The subjective perception of the participants is measured both with the SASSI (Subjective Assessment of Speech System Interfaces) tool, as well as with a custom survey for measuring the subjective perception of coherence.

\keywords{Social Robotics \and Conversational Agents \and Knowledge-Grounded Conversation}
\end{abstract}

\section{Introduction}
\label{introduction}
Social Robotics is a research field aimed at providing robots with a brand new set of skills, specifically related to social behaviour and natural interaction with humans.

Social robots can be used in many contexts such as education \cite{sar-education}, welcoming guests in hotels \cite{sar-hotels}, cruise ships \cite{sar-cruise}, malls \cite{sar-malls}, and elderly care \cite{sar-elderly}.
A noteworthy application of Socially Assistive Robots (SARs) is in the healthcare field: it has been argued that robots can be used to make people feel less lonely and help human caregivers taking care of elders in care homes \cite{caresses-protocol}\cite{caresses-tests}.
In particular, robots may help to cope with caregiver burden, i.e., the stress perceived by formal and informal caregivers, a relevant problem both in care homes and domestic environments. This subjective burden is one of the most important predictors for adverse outcomes of the care situation for the caregivers themselves, as well as for the one who requires care. Clinicians frequently overlook the caregiver burden problem \cite{burden}.

Recently, social robots have been recognized as a very important resource by one of the most recent editorials of Science Robotics \cite{combating-covid-19}, which analyzes the potential of robots during the COVID-19 pandemic and underlines how Social Robotics is a very challenging area: social interactions require the capability of autonomously handling people's knowledge, beliefs, and emotions. In the last year, COVID-19 threatened the life of people with a higher risk for severe illness, i.e., older adults or people with certain underlying medical conditions. To slow down the spread of the virus \cite{covid}, there is the need to reduce social contacts and, as a consequence, many older adults have been left even more socially isolated than before. However, in many cases, loneliness can spring adverse psychological effects such as anxiety, psychiatric disorders, depression, and decline of cognitive functions \cite{consequences}.
In this scenario, Social Robotics and Artificial Intelligence, in general, may have a crucial role: conversational robots and virtual agents can provide social interactions, without spreading the virus. 

In many Social Robotics applications, the main focus is on the conversation. Based on its purpose, the conversation can be subdivided into:
\begin{itemize}
    \item \textit{task-oriented}: it is used to give commands to perform a task or retrieve information. Examples of task-oriented conversations can be found both during the interaction with a robot, i.e., ``Go to the kitchen", and during the interaction with a smart assistant, i.e., ``Turn off the light". In this case, the problem is to understand the semantics of what the user is saying, to provide the required service. 
    \item \textit{chit-chatting}: it has the purpose of gratifying the user in the medium to long term, keeping him/her engaged and interested. For this kind of conversation, the most relevant problem, rather than perfectly grasping the semantics of what the person says, is to be able to manage the conversation, showing the knowledge and competence on a huge number of different topics coherently and entertainingly \cite{schuetzler-2020}.
\end{itemize}

In this scenario, the CARESSES project\footnote{\url{http://caressesrobot.org/}} is the first project having the goal of designing SARs that are culturally competent, i.e., able to understand the culture, customs, and etiquette of the person they are assisting, while autonomously reconfiguring their way of acting and speaking \cite{paving}. CARESSES exploited the humanoid robot Pepper\footnote{CARESSES architecture has been recently revised to allow for a Cloud-based implementation \cite{feasibility-study}, thus enabling virtually any device equipped with a network interface to perform long-term, culture-aware conversation with the user. Currently, the robots Pepper and NAO, the pill-dispenser Pillo, Prof. Einstein, and a virtual character implemented as an Android application have been provided with the onboard functionalities to implement the instructions received by the CARESSES Cloud \cite{cloud}.} as the main robotic platform for interacting with people \cite{caresses-protocol}, in particular with older people in care homes, to make them feel less isolated and reduce the caregiver burden. Under these conditions, the capability of engaging the user to chit-chat with the system becomes of the utmost importance, as it has been shown \cite{caresses-tests} that this may have a positive impact on quality of life \cite{sf36}, negative attitude towards robots \cite{nars}, and loneliness \cite{uls8}. However, for a proper conversation to be possible, the system shall be able to talk about a huge number of topics that may be more or less relevant for different cultures, by properly managing the flow of the conversation to provide coherent replies, even when it is not able to fully understand what the user is talking about (which is likely to happen very often if the user is free to raise any topic). 




To increase the impact of the conversational systems of Social Robots, \cite{mavridis} argues that they shall be designed to overcome a number of limitations. Some desirable features that are not present in most systems are: 
(1) Breaking the ``simple commands only" barrier; (2) Multiple speech acts; (3) Mixed initiative dialogue;(4) Situated language and the symbol grounding problem; (5) Affective interaction; (6) Motor correlates and Non-Verbal Communication; (7) Purposeful speech and planning; (8) Multi-level learning; (9) Utilization of online resources and services.

The main contribution of the article is to investigate possible solutions to issue 7, that we aim to achieve by properly managing the conversation flow towards the execution of tasks or the exploration of relevant topics, thus
ultimately leading to a more engaging interaction with the user.
During the conversation, it typically happens that 
a robot may provide answers that have nothing to do with what the user says: this has a very negative impact on the ``suspension of disbelief" that is required to give the user the impression of genuine intelligence, and ultimately it generates frustration. Here, a key element is the capability to know when to further explore the current topic or choose the next conversation topic, coherently with the user's sentence: if the algorithm does not pick a topic coherently, the agent's reply will not be appropriate, even if the system had the knowledge to interact consistently.


Specifically, we proceed as follows.

First, we propose a novel system for knowledge-based conversation and Dialogue Management that relies on an Ontology for the description of all relevant concepts that may play a key role in the conversation: the Ontology is designed to take into account the possible cultural differences between different users in a no-stereotyped way, and it stores chunks of sentences that can be composed in run-time, therefore, enabling the system to talk about the aforementioned concepts in a culture-aware and engaging way \cite{feasibility-study}\cite{knowledgeRepresentation}.

Second, we test the developed solutions by comparing five Artificial Conversational Agents during a conversation with 100 recruited participants: (i) an agent addressing dialogue flow management based only on the detection of keywords, (ii) an agent addressing dialogue flow management based both on the detection of keywords and the Content Classification feature of Google Cloud Natural Language, (iii) an agent that picks random topics among those present in the Ontology, (iv) a human pretending to be a chatbot, and (v) one of the most famous chatbots worldwide: Replika.
The subjective perception of the participants is measured both with the SASSI (Subjective Assessment of Speech System Interfaces) tool \cite{Lewis2015}\cite{sassi} and a custom survey for measuring the subjective perception of coherence in the conversation. 

The remainder of this article is organized as follows. Section \ref{sec:SOTA} presents an overview of previous works related to Knowledge-Grounded Conversation, the most popular validated tools to evaluate the User Experience, and introduces the concept of ``coherence" in Dialogue Management. Eventually, it presents a typical classification for Artificial Conversational Agents and describes the up-to-date most famous chatbots. An overview of the CARESSES knowledge-based conversational system is given in Section \ref{sec:system-architecture}. Section \ref{sec:materials-methods} describes the experiment carried on to evaluate the user satisfaction when interacting with the agents, along with the statistical tools used to analyse the collected data. Section \ref{sec:results} presents the results obtained, discussed in detail in Section \ref{sec:discussion}. Eventually, Section \ref{sec:conclusion} draws the conclusions.

\section{State of the Art}
\label{sec:SOTA}
This section addresses Knowledge-Grounded Conversation, presenting a brief analysis of the related Literature. Moreover, it gives an overview of the most used tools to evaluate the User Experience during the interaction with conversational agents and introduces a new metric exploited in our experiment. Eventually, it presents a classification for Artificial Conversational Agents and describes the up-to-date most famous chatbots.

\subsection{Knowledge-Grounded Conversation}
\label{know-gr}
A knowledge-grounded conversational system is a dialogue system that can communicate by recalling internal and external knowledge, similarly to how humans do, typically to increase the engagement of the user during chit-chatting. The internal knowledge is composed of things that the system already knows, while external knowledge refers to the knowledge acquired in run-time. 
To this end, knowledge-grounded systems must not only understand what the user says but also store external knowledge during the conversations, and their responses should be based on the stored knowledge. 
There are many differences between the replies of a normal conversational system and those of a knowledge-grounded system. For example, if the user says ``I love pizza", the former might provide a general answer such as ``I see, very interesting" as it has no previous knowledge about \texttt{Pizza}. On the other side, the latter may be more specific, i.e., ``Oh, pizza is a very delicious Italian food", if it had the chance to recall internal or external knowledge about pizza, e.g., using resources on the web.

Building a knowledge-grounded conversational system raises many challenges. For instance, the internal knowledge may be static or ``expandable", i.e., updated in run-time with new external information retrieved from the user utterances, which will be considered as internal knowledge in the next interactions. The external knowledge in most cases will come from websites: however, even if the system can find some documents associated with the current conversation topic by using a state-of-the-art information retrieval system, it may be difficult to extract knowledge from the search results because this would typically require complex Natural Language Processing techniques to select the appropriate knowledge.

This is done in \cite{know-pow} using a Generative Transformer Memory Network and in \cite{know-extraction} using an SVM classifier, to name a few. Another complex task is to generate appropriate responses that reflect the acquired knowledge as a consequence of the conversation history. The problems of knowledge extraction and response generation, among the others, are addressed in \cite{know-ground-chat}, where a knowledge-grounded multi-turn chatbot model is proposed (see also \ref{sec:artificial-conv-agents} for popular chatbots based on this principle).

A data-driven and knowledge-grounded conversation model, where both conversation history and relevant facts are fed into a neural architecture that features distinct encoders for the two entries, is proposed in  \cite{knowledge-grounded}. Such model aims to produce more appropriate responses: the article shows that the outputs generated by a model trained with sentences and facts related to the conversation history were evaluated by human judges as remarkably more informative (i.e., knowledgeable, helpful, specific) with respect to those of a model trained with no facts.

In \cite{social-media}, a large corpus of status-response pairs found on Twitter has been employed to develop a system based on phrase-based Statistical Machine Translation, able to respond to Twitter status posts. As stated in the article, data-driven response generation will provide an important breakthrough in the conversational capabilities of a system. Authors claim that when this kind of approach is used inside a broad dialogue system and is combined with the dialogue state, it can generate locally coherent, purposeful, and more natural dialogue.

Finally, the problem of knowledge representation and knowledge-based chit-chatting, keeping into account the cultural identity of the person, is addressed in \cite{knowledgeRepresentation} and \cite{feasibility-study} using an Ontology designed by cultural experts coupled with a Bayesian network, to avoid rigid representations of cultures that may lead to stereotypes. Such an approach allows the system to have more control over what the robot says during the conversation: the idea of having a knowledge base designed by experts is particularly suitable for sensitive situations, i.e. when dealing with the more fragile population such as older adults or children. Approaches that automatically acquire knowledge from the Internet may not be optimal in these scenarios.

\subsection{User Experience}
In the Literature, tools exist that can be used to evaluate the User Experience (UX) and the overall quality of the conversation. 
The work carried on in \cite{questionnaires-comparison} reviews the six main questionnaires for evaluating conversational systems: AttrakDiff, SASSI, SUISQ, MOS-X, PARADISE, and SUS. Moreover, it assesses the potential suitability of these questionnaires to measure various UX dimensions.

As a measure of the quality of the flow of conversation, we aim to evaluate how ``coherent" the system is: this shall be done not only taking into account the last user's utterance but also the current conversation topic (i.e., what is referred to as context in \cite{meena}\cite{blender}). To clarify what this means, suppose a conversational system that replies very accurately to everything the user says: if the user talks about football, the system will reply talking about football, if the user talks about apples, the system will reply talking about apples. But what if the person and the system are talking about football, and the person says something that is not recognized as related to anything specific, such as ``I will think about that", ``I think so", ``It's so nice to talk with you", and so on? A coherent Dialogue Management system shall be able to understand when it is more appropriate to further explore the current conversation topic (in this example, by taking the initiative to ask the person a question about his/her preferred team or players) or lead the dialogue to another topic, coherently with what the user said.

Unfortunately, the aforementioned UX tools are not meant to evaluate a medium-term mixed-initiative dialogue, as in our case. The most similar measure to what we want to evaluate is the \textit{System Response Accuracy} scale of the SASSI questionnaire: such a questionnaire, described more in detail in Section \ref{sassi}, has been used during our experiments. The \textit{System Response Accuracy} is defined as the system’s ability to correctly recognise the speech input, correctly interpret the meaning of the utterance, and then act appropriately. However, according to the aforementioned definition, what the System Response Accuracy scale aims to measure is not equivalent to the broader concept of coherence that has been given: it measures the quality of the system's reply to the user's input in a purely reactive fashion, without relying on the concept of ``current topic of conversation".

For this reason, a coherence measure in the spirit of \cite{coherence2} and \cite{coherence3} has been used to supplement the SASSI questionnaire, which requires users to rate individual sentences pronounced by the robot at the light of the context (details are given in Section \ref{coherence}). 
Notice that \cite{coherence3} and \cite{coherence1} propose also methods to automatically evaluate the coherence, as this is considered an important metrics to evaluate multi-turn conversation in open domains: however, in this work we are only interested in evaluating the subjective perception of users, which makes a simple rating mechanism perfectly fitting our purposes.

\subsection{Artificial Conversational Agents}
\label{sec:artificial-conv-agents}


Depending on the emphasis they put on a task-oriented conversation or chit-chatting, a typical classification for artificial conversational agents is based on their scope: 

\begin{itemize}
    \item \textit{Question answering bots}: knowledge-based conversational systems that answer to users’ queries by analysing the underlying information collected from various sources like Wikipedia, DailyMail, Allen AI science and Quiz Bowl \cite{chatbot-types};
    \item \textit{Task-oriented bots}: conversational systems that assist in achieving a particular task or attempt to solve a specific problem such as a flight booking or hotel reservation \cite{customer-care};
    \item \textit{Social bots}: conversational systems that communicate with users as companions, and possibly entertain or give recommendations to them \cite{customer-service}: recent notable examples are Microsoft Xiaoice \cite{xiaoice-design} and Replika\footnote{\url{https://help.replika.ai/hc/en-us/articles/115001070951-What-is-Replika-}}.
\end{itemize}


In the following, we discuss more in detail the agents belonging to the third class pointing out differences and similarities with out solution, without making a distinction between robots and chatbots, unless strictly required.


For many decades, the development of social bots, or intelligent dialogue systems that can engage in empathetic conversations with humans, is one of the main goals of Artificial Intelligence. As stated in \cite{eliza-xiaoice}, early conversational systems such as Eliza \cite{eliza}, Parry \cite{parry}, and Alice \cite{alice} were
designed to mimic human behaviour in a text-based
conversation, hence to pass the Turing Test \cite{turing} within a controlled scope. Despite their impressive successes, these systems, which were precursors to today’s social chatbots, worked well only in constrained environments. 
In more recent times, among the most successful conversational agents for general use that can act as digital friends and entertainers, Xiaoice, Replika, Mitsuku, and Insomnobot-3000 are the most frequently mentioned \cite{top-chatbots}.

\textit{XiaoIce} (``Little Ice" in Chinese) is one of the most popular chatbots in the world. It is available in 5 countries (i.e., China, Japan, US, India, and Indonesia) under different names (e.g., Rinna in Japan) on more than 40 platforms, including WeChat, QQ, Weibo, and Meipai in China, Facebook Messenger in the United States and India, and LINE in Japan and Indonesia. 
Its primary goal is to be an AI companion with which users form long-term emotional connections: this distinguishes XiaoIce not only from early chatbots but also from other recently developed conversational AI personal assistants such as Apple Siri, Amazon Alexa, Google Assistant, and Microsoft Cortana. 

As stated in \cite{xiaoice-design}, the topic database of XiaoIce is periodically updated by collecting popular topics and related comments and discussions from high-quality Internet forums, such as Instagram in the US and the website douban.com in China. To generate responses, XiaoIce has a paired database that consists of query-response pairs collected from two data sources: human conversational data from the Internet, (e.g., social networks, public forums, news comments, etc.), and human-machine conversations generated by XiaoIce and her users. 
Even if the data collected from the Internet are subjected to quality control to remove personally identifiable information (PII), messy code, inappropriate content, spelling mistakes, etc., the knowledge acquisition process is automated and it is not supervised by humans: differently from our system, this solution may not be suitable in sensitive situations, e.g., when dealing with the more fragile population such as older people or children. Regarding issue 7 in Section \ref{introduction} (i.e., the lack of flow in the conversation), the implementation of XiaoIce addressed this problem by using a Topic Manager that mimics human behavior of changing topics during a conversation. It consists of a classifier for deciding at each dialogue turn whether or not to switch topics and a topic recommendation engine for suggesting a new topic. Topic switching is triggered if XiaoIce does not have sufficient knowledge about the topic to engage in a meaningful conversation, an approach that our system adopts as well by relying on an Ontology for representing relationships among different topics of conversation. Unfortunately, we could not try XiaoIce as it is not available in Italy. 

\textit{Replika} is presented as a messaging app where users answer questions to build a digital library of information about themselves.
Its creator, a San Francisco-based startup called Luka, sees 
see a whole bunch of possible uses for it: a digital twin to serve as a companion for the lonely, a living memorial of the dead, created for those left behind, or even, one day, a version of ourselves that can carry out all the mundane tasks that we humans have to do, but never want to.

To the best of the authors' knowledge, there is no detailed explanation of the implementation of Replika, hence we can only make assumptions on how issues related to Dialogue Management are faced. After having intensively tried Replika, it seems that the conversation is partitioned into ``sessions" in which it has specific competencies, somehow playing a similar role as topics in XiaoIce and our system. Concerning issue 7 in Section \ref{introduction}, the chatbot uses a Neural Network to hold an ongoing, one-on-one conversation with its users, and over time, learn how
to talk back \cite{wired}. The agent is trained on texts from more than 8 million web pages, from Twitter posts to Reddit forums, and it can respond in a thoughtful and human-like way. From time to time, it will push the user to have a ``session” together: in this session, it will ask questions regarding what the user did during the day, what was the best part of the day, what is the person looking forward to tomorrow, and eventually, the user has to rate his/her mood on a scale of 1 to 10. During this session, Replika takes control of the conversation and insists that the user answers its questions about a specific matter, something that may be annoying and frustrating, and we avoid by letting the user free to easily switch to another topic at any time.
Replika's responses are backed by Open AI's  GPT-2\footnote{\url{https://en.wikipedia.org/wiki/GPT-2}} text-generating AI system \cite{cnn-replika}. 

\textit{Mitsuku}\footnote{\url{https://www.pandorabots.com/mitsuku/}}, or Kuki as her close friends call her, is a chatbot created by Pandorabots\footnote{\url{https://home.pandorabots.com/home.html}}: an open-source chatbot framework that allows people to build and publish AI-powered chatbots on the web, mobile applications, and messaging apps like LINE, Slack, WhatsApp, and Telegram. 

The Pandorabots chatbot framework is based on the Artificial Intelligence Markup Language (AIML) scripting language, which developers can use to create conversational bots. The downside is that Pandorabots do not include machine learning tools that are common on other chatbot building platforms. A specific AIML file allows users to teach Mitsuku new facts: the user should say ``Learn" followed by the fact (i.e., ``Learn the sun is hot"). The taught information is emailed to its creator Steve Worswick, which will personally supervise the learning process. To address issue 7 in Section \ref{introduction}, specific areas of competence of the chatbot are managed by different AIML files, which are responsible for maintaining coherence in the responses: AIML files, therefore, play a similar role as topics/concepts in the Ontology that are the core elements of our system, but without the benefit of having a hierarchical structure as a key element for Dialogue Management.
Mitsuku is a five-time Loebner Prize winner (in 2013, 2016, 2017, 2018, 2019), and it is available in a web version\footnote{\url{https://chat.kuki.ai}}, Facebook Messenger, Kik Messenger, and Telegram.  

\textit{Insomnobot-3000}\footnote{\url{https://insomnobot3000.com}} is the world's first bot that is only available to chat, exclusively via SMS, between 11pm and 5am regardless of the time zone. The bot was built by the mattress company Casper and, according to its creators, it was programmed to sound like a real person and talk about almost anything. Insomnobot-3000 cannot learn new things and expand its knowledge base: however, it can generate over 2,000 different responses, depending on which category and emotion the keyword falls under. Concerning issue 7 in Section \ref{introduction}, when the bot receives a text, it chooses an appropriate response by identifying keywords: however, differently from the aforementioned systems and our solution, it lacks a more sophisticated mechanism for Dialogue Management to switch in-between different areas of competence or topics to provide contextual replies. 

In addition to the aforementioned chatbot available to the large public, recent research on data-driven knowledge-grounded conversation based on sophisticated generative models is worth being mentioned.

\textit{Meena} \cite{meena} is a generative chatbot model trained end-to-end on 40B words mined from public domain social media conversations, that addresses the problem of multi-turn conversation in open domains.
According to its authors, this chatbot can conduct conversations that are more sensible and specific than existing state-of-the-art chatbots. Such improvements are measured by a new human evaluation metric, called Sensibleness and Specificity Average (SSA), which captures important attributes for human conversation. Meena is based on the concept of multi-turn contexts to evaluate the quality of a response: not only the last sentences but the recent conversation history in terms of \textit{(context, response)} pairs is used to train and evaluate the network. Experiments show significant improvements in SSA with respect to competitors, providing replies that are coherent with what the user said and, at the same time, more specific to the context. Authors report also that issues related to inappropriate and biased language have still to be solved: to the best of our knowledge, a publicly available version of this chatbot has not been released yet.

\textit{BlenderBot} \cite{blender} is a very recent open-domain chatbot developed at Facebook AI Research. 
As like as we do, the authors put a stronger emphasis on aspects that are not only related to sentence generation: they argue that good conversation requires a number of skills to be blended in a seamless way, including providing engaging topics of conversation as well as listening and showing interest to what the user says, among the others. Current generative models, they argue, tend to produce dull and repetitive responses (and, we would add, they do not take cultural appropriateness into account), whereas retrieval models may produce human written utterances that tend to include more vibrant (and culturally appropriate) language. To overcome the limitations of purely generative models, BlenderBot includes a retrieval step before generation according to a retrieve and refine model \cite{Weston}, which retrieves initial dialogue utterances and/or knowledge from a large knowledge base. As like Meena, the system is trained to choose the next dialogue utterance given the recent dialogue history, referred to as context: training is performed using a number of datasets that are key to exhibit different skills (i.e., engaging personality, emotional talking, knowledge-grounded conversation), as well as a Blended Skill Talk dataset that merges utterances taken from the three. Authors claim that their model outperforms Meena in human evaluations, however, they acknowledge that the system still suffers from a lack of in-depth knowledge if sufficiently interrogated, a tendency to stick to simpler language, and a tendency to repeat often used phrases.

DialoGPT \cite{dialogpt} is a neural conversational response generation model trained on 147M conversation-like exchanges extracted from Reddit comment chains over a period spanning from 2005 through 2017 and it extends GPT-2. Authors claim that conversational systems that leverage DialoGPT generate more relevant, contentful, and context-consistent responses than strong baseline systems.
As it is fully open-sourced and easy to deploy, users can extend the pre-trained conversational system to bootstrap training using various datasets. Authors claim that the detection and control of toxic output will be a major focus of future investigation.


All generative models based on human-human data offers many advantages in producing human-like utterances fitting the context, but have the major drawback that they can learn undesirable features leading to toxic or biased language. Classifiers to filter out inappropriate language exist \cite{toxic}, but they still have limitations: this issue is particularly important when dealing with the more frail populations, especially by considering that classifying language as inappropriate or offensive typically depends on cultural factors, and the development of classifiers working properly in multiple cultures may be very challenging. Since we claim that a proper flow of conversation deserves higher attention than the automatic generation of sentences, we address these problems through an Ontology of conversation topics and topic-related chunks of sentences generated by cultural experts, that are then composed in run-time using the hierarchical structure of the knowledge-base to produce a huge variety of sentences. Finally, even if this is not the main objective of this article, it is worth reminding that our system may take cultural aspects into account in Dialogue Management, an element that is completely ignored by all state-of-the-art systems.


\section{System Architecture}
\label{sec:system-architecture}
Even if the focus of this article is on knowledge-based chit-chatting and dialogue flow management, and not on cultural adaptation, it is necessary to briefly introduce the structure of the cultural knowledge base used by the CARESSES conversational system since it strongly influences the algorithms we implemented. 

\subsection{Knowledge Representation}
\label{knowledge-representation}
In CARESSES, the ability of the companion robot to naturally converse with the user has been achieved by creating a framework for cultural knowledge representation that relies on an Ontology \cite{ontology} implemented in OWL2 \cite{owl2}. According to the Description Logics formalism,  concepts (i.e., topics of conversation the system is capable of talking about) and their mutual relations are stored in the terminological box (TBox) of the Ontology. Instead, instances of concepts and their associated data (e.g., chunks of sentences automatically composed to enable the system to talk about the corresponding topics) are stored in the assertional box (ABox). 

\begin{figure}[ht]
    \centering
    \includegraphics[width=\linewidth]{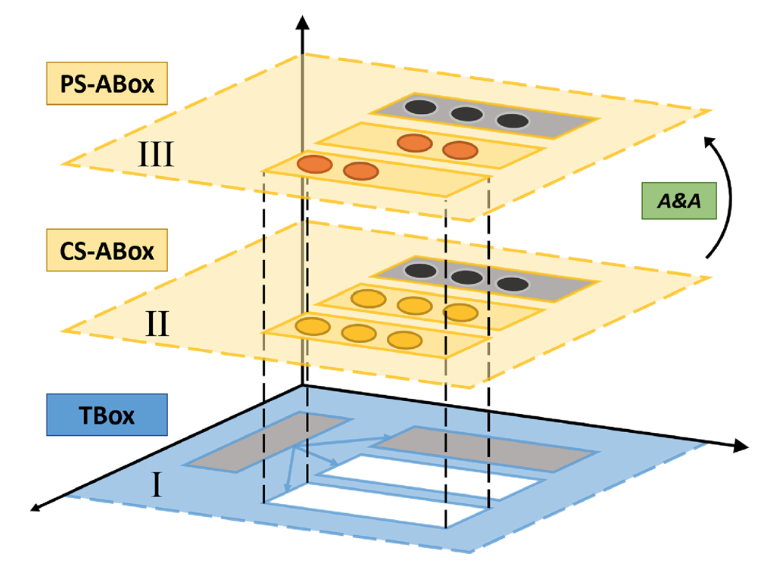}
    \caption{Knowledge representation architecture for a culturally competent robot.}
    \label{fig:three-layer-Ontology}
\end{figure}

\begin{figure}
    \centering
    \includegraphics[width=\linewidth]{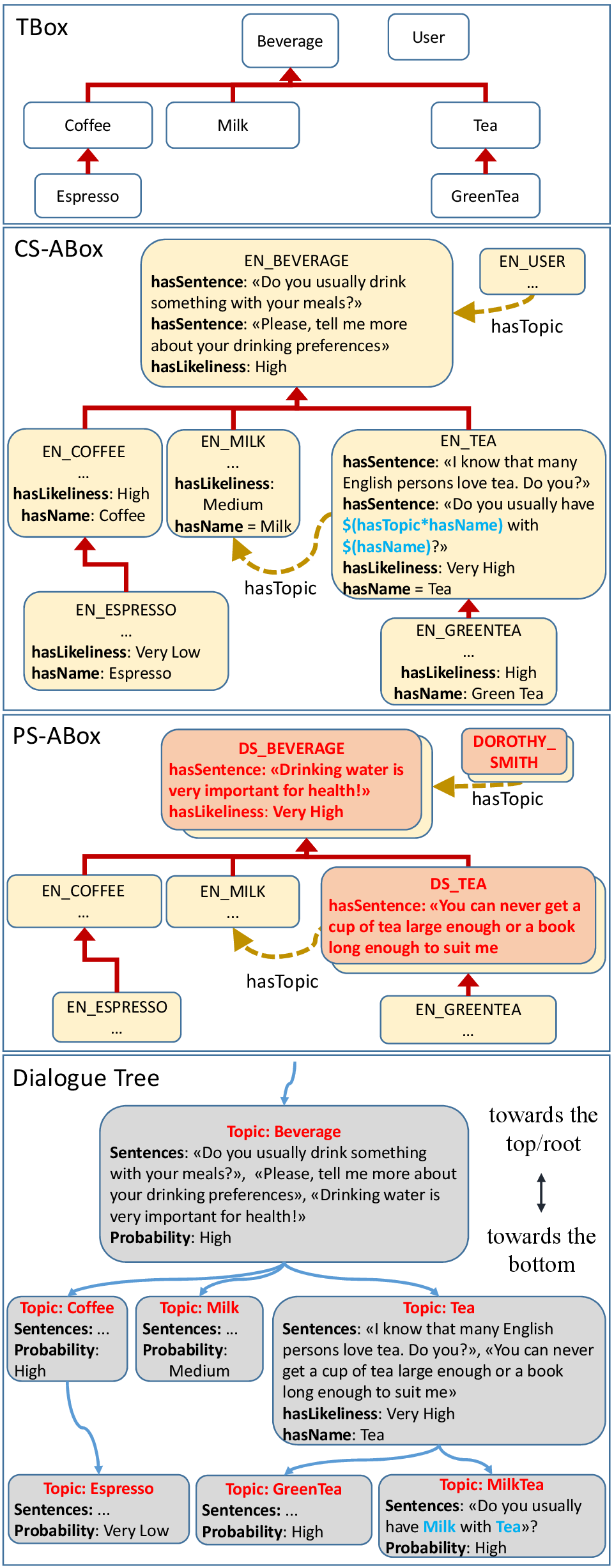}
    \caption{The three layers of the Ontology: TBox, CS-ABox (for the English culture), PS-ABox (for the user Dorothy Smith), and the Dialogue Tree generated from the Ontology structure.}\label{fig:architecture}
\end{figure}

To deal with representations of the world that may vary across different cultures \cite{Carrithers2010}, the Ontology is organized into three layers, as shown in Figure \ref{fig:three-layer-Ontology}. 
The TBox (layer I) encodes concepts at a generic, culture-agnostic level, which can be inherited from existing upper and domain-specific Ontologies (grey boxes) or explicitly defined to enable culture-aware conversation with the user (white boxes). An important point to be highlighted is that the TBox should include concepts that are typical of the union all cultures considered, whichever the cultural identity of the user is, to avoid stereotypes (an example related to different kind of beverages is shown in Figure \ref{fig:architecture}). The system will initially guess the user's beliefs, values, habits, customs, preferences based on the culture he/she declared to self-identify with, but it will then be open to considering choices that may be less likely in a given culture, as the user explicitly declares his/her attitude towards them. According to this principle, the system may initially infer that an English person may be more interested to talk about \texttt{Tea} rather than \texttt{Coffee}, and the opposite may be initially inferred for an Italian user. However, during the conversation, initial assumptions may be revised, thus finally leading to a fully personalized representation of the user's attitude towards all concepts in the TBox, to be used for conversation.

To implement this mechanism, the Culture-Specific ABox layer comprises instances of concepts (with prefix \texttt{EN\_} for ``English" in Figure \ref{fig:architecture}) encoding culturally appropriate chunks of sentences to be automatically composed (Data Property \texttt{hasSentence}) and the probability that the user would have a positive attitude toward that concept, given that he/she belongs to that cultural group (Data Property \texttt{hasLikeliness}). 

Eventually, the Person-Specific ABox comprises instances of concepts (with prefix \texttt{DS\_} for a user called ``Dorothy Smith" in Figure \ref{fig:architecture}) encoding the actual user’s attitude towards a concept 
(
Mrs Dorothy Smith may be more familiar with having tea than the average English person, 
\texttt{hasLikeliness=Very High}), sentences to talk about a topic explicitly taught by the user to the system (\texttt{hasSentence}=``You can never get a cup of tea large enough or a book long enough to suit me") or other knowledge 
explicitly added during setup (e.g., the user's name and the town of residence). At the first encounter between the robot and a user, many instances of the Ontology will not contain Person-Specific knowledge: the robot will acquire this awareness at run-time either from its perceptual system or during the interaction with the user, e.g., asking questions. 
Figure \ref{fig:three-layer-Ontology} also shows that some instances of existing Ontologies that are not culture- or person-dependent (dark circles) may not change between the two ABox layers: this may refer, for instance, to technical data such as the serial number of the robot or physical quantities, if they need to be encoded in the Ontology. For a detailed description of the terminological box (TBox) and assertional box (ABox) of the Ontology, as well as the algorithms for cultural adaptation, see \cite{feasibility-study}\cite{knowledgeRepresentation}.

The Dialogue Tree (DT) (Figure \ref{fig:architecture}), used by the chit-chatting system (Section \ref{chit-chatting}), is built starting from the Ontology structure: each concept of the TBox and the corresponding instances of the ABox are mapped into a conversation topic, i.e., a node of the tree. The relation between topics is borrowed from the structure of the Ontology: specifically, the Object Property \texttt{hasTopic} and the hierarchical relationships among concepts and instances are analyzed to define the branches of the DT. In the example of Figure \ref{fig:architecture}, the instance of \texttt{Tea} for the English culture is connected in the DT to its child node \texttt{GreenTea} (which is a subclass of \texttt{Tea} in the Ontology), and its sibling \texttt{MilkTea} (since \texttt{EN\_MILK} is a filler of \texttt{EN\_TEA} for the Object Property \texttt{hasTopic}). 

As mentioned, each conversation topic has chunks of culturally appropriate sentences associated with it that are automatically composed and used during the conversation. Such sentences can be of different types (i.e., positive assertions, negative assertions, different kinds of questions, or proposals for activities) and are selected in subsequent iterations, also depending on input received from the user. Typically, when exploring a topic, the system may first ask a question to understand if the user is familiar with that topic: if so, this may be followed by general considerations, proposals for activities, or open questions, until the system moves to the next topic if the conditions hold. Multiple sentences of the same type are present, randomly chosen to reduce the chance of repetitions. At any time, the user is free to take the initiative to express his/her considerations or lead the conversation somewhere else.
Sentences may contain variables that are instantiated when creating the DT. For instance, a hypothetical sentence ``Do you like \$hasName?" encoded in the concept \texttt{Coffee} might be used to automatically produce both ``Do you like Coffee?" and ``Do you like Espresso?" (being \texttt{Espresso} a subclass of \texttt{Coffee}). As further example, the sentence ``I love \$hasName with \$hasTopic*hasName" encoded in the concept \texttt{Movie} can be used to produce both ``I love movies with great actors" and ``You know... I love Bollywood movies with Amitabh Bachchan", depending on the taxonomy of the knowledge base and additional rules for sentence composition. In the current version used for testing, exploiting the taxonomy of the Ontology allowed us to easily produce a DT with 2,470 topics of conversation and 24,033 sentences, with random variations made in run-time.

A few additional words are worth spending to clarify this concept. Currently, as in the aforementioned example, one sentence may contain two variables whose value depends on (i) the specific concept that inherited that Data Property from superconcepts in the Ontology, and (ii) related concepts connected to it through an Object Property. Then, when manually adding a sentence as a Data Property of a given concept, a number of sentences will be inherited and added automatically depending on the number of (i) its subconcepts and (ii) their role-fillers. By considering a DT generated from the Ontology and having a branching factor $B=B_s + B_p$, where $B_s$ is the branching factor due to subconcepts and $B_p$ is the branching factor due to role-fillers, adding a sentence in a concept at a height $H$ from DT leaves produces a number of sentence in the order of $O(H^{B})$; by considering that all sentences are randomly associated in run-time with a prefix (e.g., ``You know...", ``I heard that...") and may be appended to each other, this increases the number of variations even more. For instance, in the current configuration with 2,470 concepts, adding a sentence at the root of the DT will produce a maximum of 2,470 inherited sentences in the whole Ontology (which however, may lack specificity, as it is unlikely that the same pattern may be easily adapted to talk about everything in the ontology: i.e., ``I love talking about \$hasTopic with you" may work for all subconcepts, but ``I like eating \$hasTopic" will not).

\subsection{Chit-Chatting}
\label{chit-chatting}
When the user pronounces a sentence that triggers the start of a conversation, the system switches to the \textit{chit-chatting} state. In this case, the Dialogue Management algorithm operates to keep the conversation as engaging and natural as possible by implementing two different, intertwined methods to navigate the DT based on what the user says:

\begin{enumerate}
    \item The first mechanism implemented by the Dialogue Management algorithm simply chooses the topic based on the DT, following the branches from more general conversation topics close to the root (e.g., the topic \texttt{BreakfastHabits}) to more specific (e.g., the topic \texttt{HavingTeaForBreakfast}) or related ones (e.g., the topic \texttt{TakingCareOfOneself}). This is achieved by selecting the branches to maximise the probability that the user will be interested in the next topic (also based on cultural factors). When in a topic, the system asks questions, reply with positive or negative assertions, propose activities or simply encourage the user to freely express his/her view, by composing chunks of sentences encoded in the ontology, until the topic has been sufficiently explored.
    Probabilities associated with topics have been initially assigned with the help of Transcultural Nursing experts as well as looking for information on the Internet about habits, foods, sports, religions, etc. in different cultures. Please, notice however that these are only initial guesses, as the system will update probabilities during the conversation depending on the user's preferences and attitudes to avoid stereotyped representations.
    \item The second mechanism enables the Dialogue Management algorithm to jump to a different topic of the DT depending on what the user says (e.g., when talking about \texttt{BreakfastHabits} the person may start talking about a \texttt{Restaurant} he/she loves). After doing so, the system reverts to the first mechanism but starting from the new topic: it asks questions, makes positive and negative comments, proposes activities, or allows the person to freely express his/her view about the new topic, and then re-start following branches as previously.
\end{enumerate}

With reference to Figure \ref{fig:architecture}, a possible dialogue pattern between the robot and Mrs Dorothy Smith, using the aforementioned mechanisms, would be the following (labels 1 and 2 identify the mechanism currently adopted by the system to navigate the DT):

\textit{R. (1): Do you usually drink something with your meals?}

\textit{Mrs S.: Always! At least I try...}

\textit{R. (1): Drinking water is very important for health! Please, tell me more about your drinking preferences.}

\textit{Mrs S.: Well, I mostly drink water during lunch.}

\textit{R. (1): I know that many English persons love tea. Do you?}

\textit{Mrs S.: Yes, of course! But mostly in the afternoon.}

\textit{R. (1): You can never get a cup of tea large enough or a book long enough to suit me. Do you usually have milk with your tea?}

\textit{Mrs S.: Yes, and scones, too! But they are not very healthy.}

\textit{R. (2): I love scones with jam and whipped cream!}

\begin{figure}
    \centering
    \includegraphics[width=\linewidth]{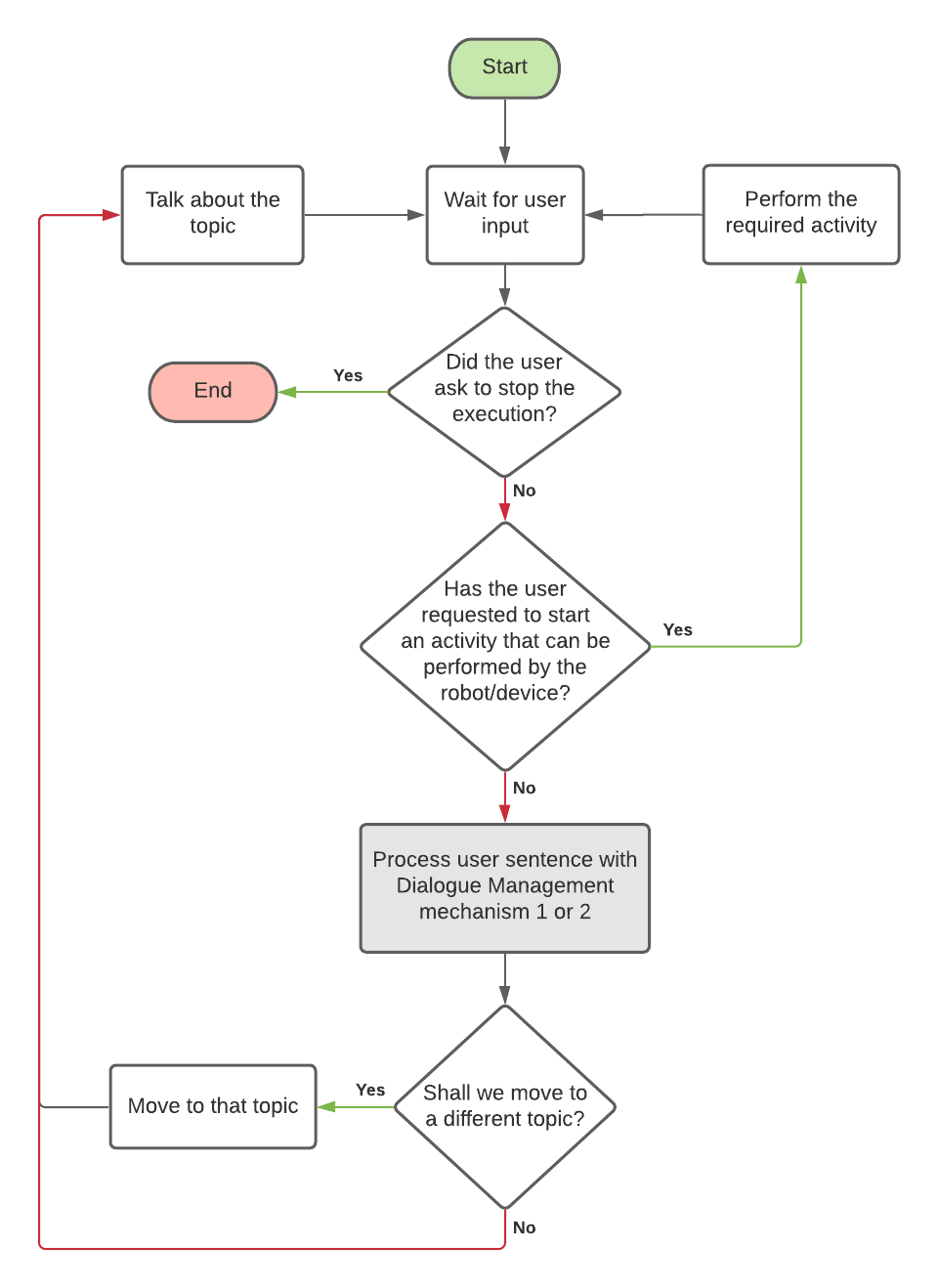}
    \caption{Flow chart of the basic chit-chatting framework.}
    \label{fig:initial-diagram}
\end{figure}

Figure \ref{fig:initial-diagram} depicts a detailed flow-chart that clarifies what happens during the interaction with the user. 

The diagram shows that, as soon as the person says something, the Dialogue Management algorithm first checks if the user's sentence contains some command that is aimed to trigger an activity (box \texttt{Perform the required activity}). This task-oriented part of the dialogue has been implemented in our experiments with caring robots during the CARESSES project, but not in the tests described in this article, and it may rely on a third-party system for NLP\footnote{For instance, \url{https://dialogflow.cloud.google.com/}, that in the current implementation also manages small-talk requests triggered by sentences such as ``Hello", ``How are you", or similar).} or a pre-existing command interface that is already implemented onboard the robot (e.g., verbal commands to clean the room in case of a robotic vacuum cleaner, or to deliver medicines in case of a robotic pill dispenser). 

If a command is not found, the user's input is processed (box \texttt{Process user sentence with Dialogue Management mechanism 1 or 2}) by using the two mechanisms described so far: as a first step the user's utterance is analysed to check if something relevant is detected, and possibly used to jump to another topic using mechanism 2; if no information is found to jump to another topic, mechanism 1 checks if the current topic shall be further explored or it is time to follow the branches of the DT from a parent node to a children node that is semantically related in the Ontology. Eventually, it should be mentioned that activities (e.g., vacuum cleaning or pill delivery) may even be proposed by the Dialogue Management System itself when related to the current topic of discussion (i.e., if the person is talking about cleaning and the robot has this capability). 
The interaction continues in this way until the user explicitly asks to stop it.

This approach to Dialogue Management, despite its simplicity, reveals to be very effective for chit-chatting. It should be reminded that here Dialogue Management is not meant to understand all the nuances of human language when giving a specific command to be executed (i.e., ``Put the red ball on the green table"), but rather it aims at engaging the user in rewarding conversations: an objective that we try to achieve by enabling the system to coherently talk about a huge number of different topics while properly managing the conversation flow.
Please notice that the approach has been already tested in the CARESSES project with more than 20 older people of different nationalities, for about 18 hours of conversation split into 6 sessions: both the quantitative and qualitative analysis performed revealed very positive feedback from the participants \cite{caresses-protocol}\cite{caresses-tests}.  
However, the quantitative analysis performed with care home residents was not explicitly aimed at evaluating the quality of the conversation: the analysis performed was rather aimed to measure improvement in health-related quality of life (SF-36 \cite{sf36}), negative attitude towards robots (NARS \cite{nars}), loneliness (ULS-8 \cite{uls8}).
Indeed, additional analysis with a wider population is needed to evaluate the quality of the conversation produced by our system in terms of issue 7 in Section \ref{introduction}, and motivates the present study. 
When navigating in the knowledge base following the branches of the DT (aforementioned mechanism 1), coherence in the flow of conversation is straightforwardly preserved by the fact that two nodes of the DT are semantically close to each other by construction. However, to preserve coherence when deciding if it is needed to jump from one node to another (mechanism 2), proper strategies should be implemented to avoid the feeling that the system is ``going off-topic" with respect to what the person says: on the one side, the system needs to be enough responsive to promptly move to another topic if the person wants to; on the other side, the system needs to avoid jumping from one topic to another, overestimating the desire of the person to talk about something else when he/she would be happier to further explore the current topic of conversation.

\subsection{Jumping to a different discussion topic}
\label{dialogue-algorithm}


When using the two mechanisms described in the previous section to navigate the DT, to preserve coherence in the conversation flow the system uses one of the two following methods: keyword-based topic matching and keyword- and category-based topic matching. These methods differ in terms of complexity and in terms of the need to rely on third-party services to analyze the semantic content of sentences.

\subsubsection{Keyword-Based Topic Matching}
\label{old-dialogue-algorithm}
The first, and the simpler, method is exclusively based on the detection of keywords in the user sentence (keywords are manually encoded in the Ontology in a corresponding Data Property).
To match a topic, at least two keywords associated with that topic should be detected in the sentence pronounced by the user, using wildcards to enable more versatility in keyword matching. The use of multiple keywords allows the system to differentiate between semantically close topics (i.e., \texttt{Green Tea} rather than the more general concept of \texttt{Tea}).

Figure \ref{fig:keyw-based-alg} shows a possible implementation of the box \texttt{Process user sentence with Dialogue Management mechanism 1 or 2} in Figure \ref{fig:initial-diagram} when using the keyword-based topic matching method:

\begin{figure}
    \centering
    \includegraphics[width=\linewidth]{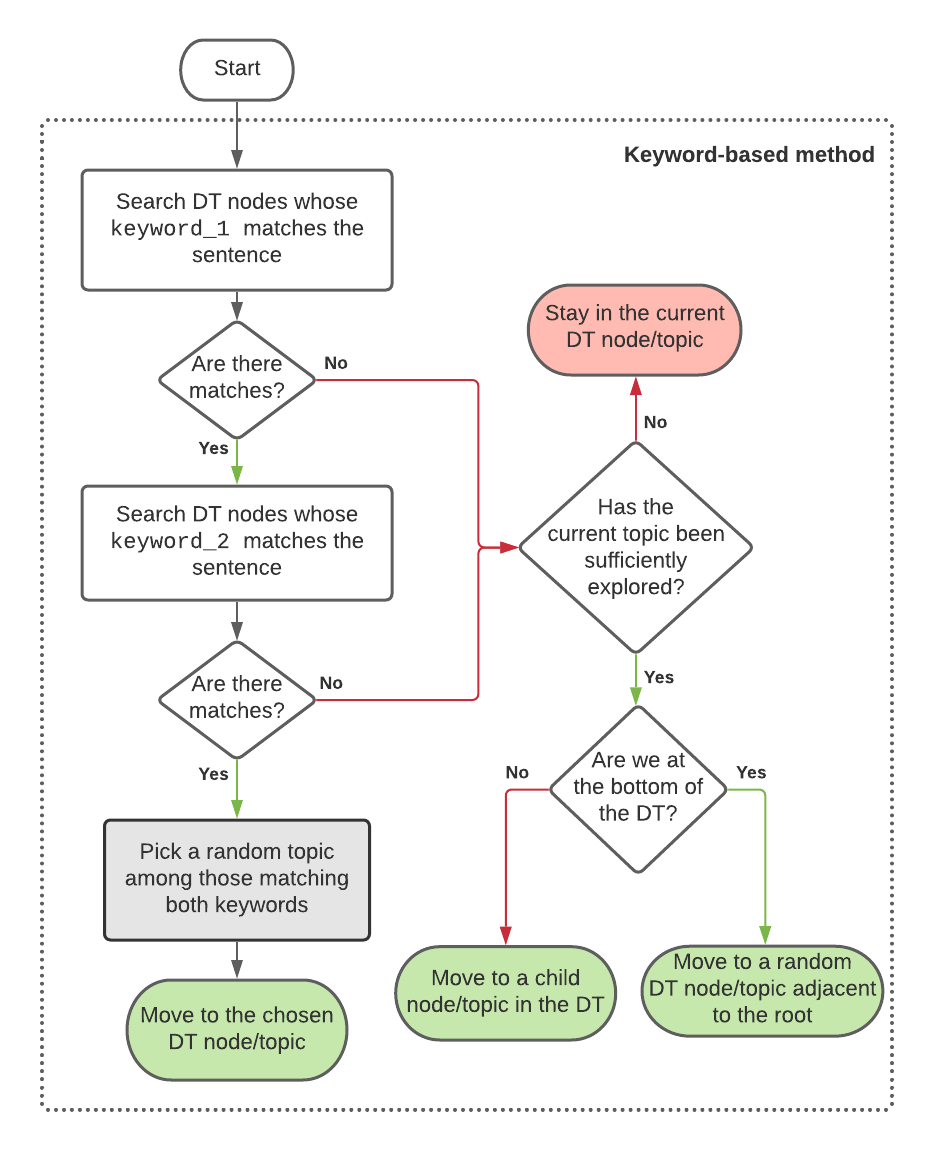}
    \caption{Flow chart of the keyword-based topic matching method.}
    \label{fig:keyw-based-alg}
\end{figure}

\begin{itemize}
    \item If the user's sentence contains keywords corresponding to a topic in the Ontology (left part of the diagram), the algorithm returns a topic of the DT that matches the keywords. In case more matches are found, the algorithm randomly chooses one of the matching topics.
    \item Otherwise:
    \begin{itemize}
        \item If there are still relevant questions to be asked or assertions to be made about that topic, the system stays on the same topic.
        \item Otherwise:
            \begin{itemize}
        \item If not at the bottom of the DT, the Dialogue Management System keeps on exploring the DT along its branches;
        \item Otherwise the algorithm returns a random topic immediately below the root.
        \end{itemize}
    \end{itemize}
\end{itemize}

This basic approach, despite its simplicity, has been successfully exploited during the experimental trial in care homes, and its capability to provide (almost always) coherent, knowledge-grounded replies, was confirmed in public exhibitions. However, when focusing on coherence, the approach has obvious limitations. 
Let's suppose that the user is having a conversation with the system and at some point, for some reason, the user says ``My bank account has a high interest". This sentence, like any other sentence, is provided as input to the keyword-based method, to find the most appropriate topic to continue the conversation. The algorithm finds the word ``interest" matching with the \texttt{keyword\_1} of the topic \texttt{HOBBY}. Then it should ideally check if the sentence contains a word matching with \texttt{keyword\_2} associated with the same topic: however, in this case, it it turns out that no second keyword is needed since a wildcard is used in the topic \texttt{HOBBY} to guarantee that one keyword is sufficient. As a result, the algorithm will return the only matching node of the DT: the system will start asking the user about his/her hobbies (which does not look very appropriate).
The second limitation related to the behaviour of this algorithm arises when no keywords are found and we are at the bottom of the DT: in this case, as already mentioned, the algorithm has no information to continue the conversation and \textit{jumps} to a random topic close to the root.

\begin{figure}
    \centering
    \includegraphics[width=\linewidth]{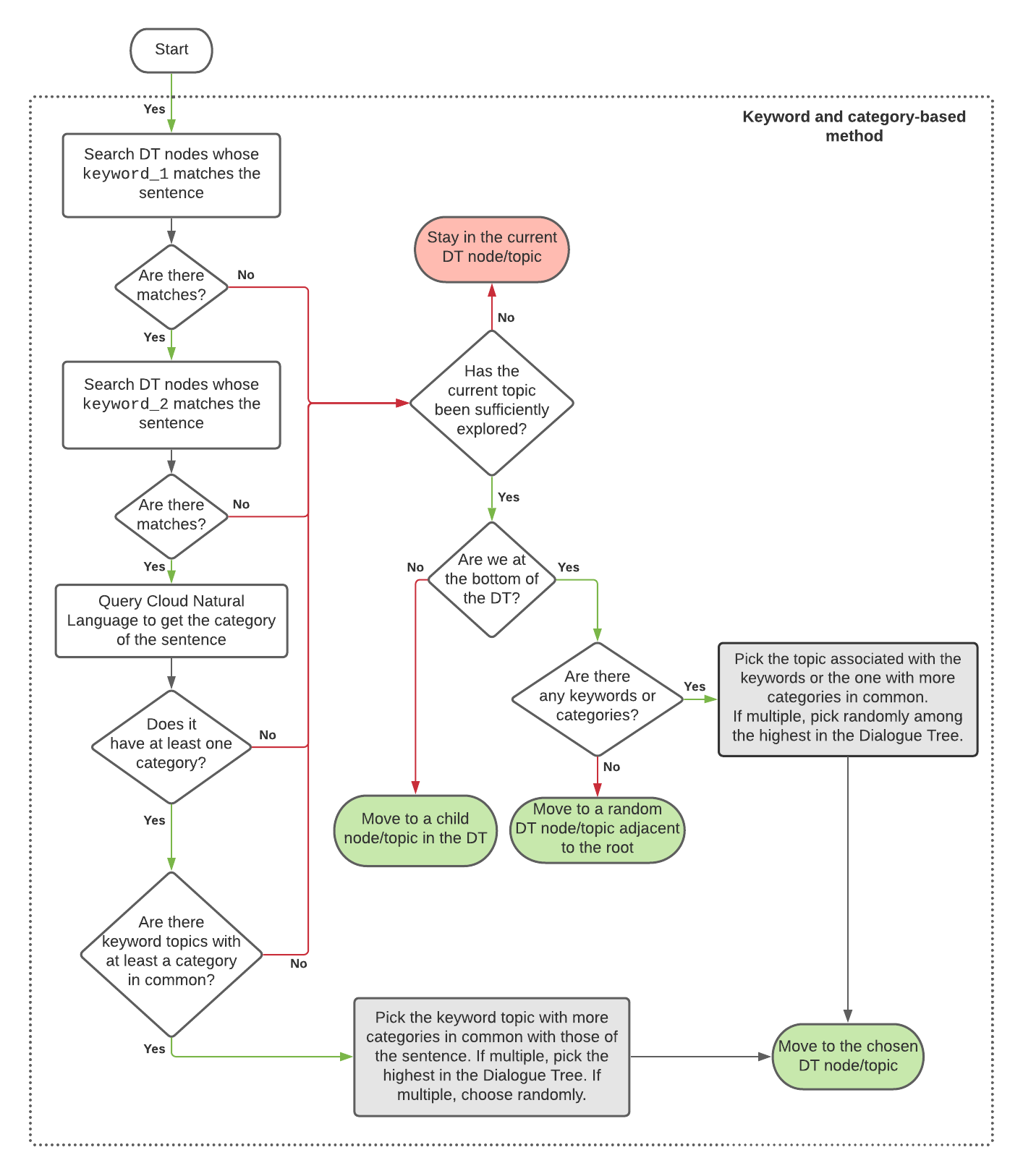}
    \caption{Flow chart of the keyword- and category-based topic matching method.}
    \label{fig:keyw-cat-based-alg}
\end{figure}

\subsubsection{Keyword- and Category-Based Topic Matching}
\label{new-dialogue-algorithm}
The second algorithm is meant to solve the issues of the previous algorithm, or at least reduce their frequency, by enabling a more complex understanding of the semantic content of the sentence. 
For this purpose, this algorithm not only exploits the keywords, but it also takes into account the ``category" of the user's sentence: in the current implementation, it uses the third-party services provided by Cloud Natural Language (CNL) API\footnote{\url{https://cloud.google.com/natural-language?hl=en}}, an advanced tool for Sentiment Analysis, Entity Recognition, and Content Classification, among the others.

To exploit the information related to the category of the sentence, the Content Classification of the topics contained in the Ontology is a fundamental operation and needs to be performed in a setup phase to create a mapping between the topics contained in the Ontology and the CNL category hierarchy.
Off-line, before the system starts, the classification procedure is performed for all the topics: the algorithm puts together all the sentences associated with each topic (inserted into the Ontology by experts and then automatically composed, or added by the users during the conversation through a mechanism not described in this article \cite{Grassi2021}), sends them to CNL, and associates the returned categories to the corresponding topic in the Ontology and then in the DT (notice that, whilst keywords are manually encoded in the Ontology, categories are automatically assigned). On-line, during the conversation, this mapping will allow the system to find which topics of the Ontology match best (i.e., have more categories in common) with the category (if any) of the sentence pronounced by the user according to CNL.
In the current Ontology, only 122 topics have not been associated with a category, as none was found to be appropriate. Hence, the topics mapped to at least a CNL Category\footnote{\url{https://cloud.google.com/natural-language/docs/categories}} are 2,348.

Figure \ref{fig:keyw-cat-based-alg} shows a possible implementation of the box \texttt{Process user sentence with Dialogue Management mechanism 1 or 2} in Figure \ref{fig:initial-diagram} when using both keywords and categories for topic matching. As a first step, the algorithm checks whether the sentence contains at least 20 tokens: the Content Classification feature of CNL does not work if the input does not satisfy this requirement. If the sentence is not long enough, it is replicated until it contains at least 20 words. Then it proceeds as follows:

\begin{itemize}
    \item If the user's sentence contains keywords corresponding to a topic in the Ontology \textit{and} CNL returns at least a category associated with the sentence (left part of the diagram), the algorithm selects the topic(s) in the Ontology with the greatest number of categories in common: if there is more than one topic, it picks the closest to the DT root or, if there are multiple topics at the same level, it chooses randomly;
    \item Otherwise, if no keywords match \textit{or} the sentence does not have any category:
    \begin{itemize}
       \item If there are still relevant questions to be asked or assertions to be made about that topic, the system stays on the same topic.
        \item Otherwise:
        \begin{itemize}
                \item If not at the bottom of the DT, the Dialogue Management System keeps on exploring the DT along its branches;
                \item Otherwise:
                \begin{itemize}
                    \item If there are keywords \textit{or} CNL categories associated with the sentence (we already know there are not both), the algorithm selects the topic(s) with matching keywords \textit{or} the greatest number of categories in common: if multiple, it picks the one closer to the root in the DT or randomly if they are at the same level.
                    \item Otherwise, the algorithm returns a random topic immediately below the root.
                \end{itemize}
        \end{itemize}
    \end{itemize}
\end{itemize}

This second method is more complex and computationally expensive than the previous one (and, in the current implementation, it requires third-party services), but it provides more stability and consistency when selecting the next conversation topic. Also, it diminishes the chances of performing random changes of conversation topic when at the bottom of the DT and no keywords are found: if the sentence has at least one category associated, recognized by CNL, there is still a chance of finding a coherent topic to jump to in order to continue the conversation.

\section{Materials and Methods}
\label{sec:materials-methods}
Experimental tests with recruited participants have been performed for multiple purposes:
\begin{itemize}
    \item Evaluate the impact of different solutions for Dialogue Management we developed, either keyword-based or both keyword- and category-based, to improve the subjective perception of the system in terms of coherence and user satisfaction; 
    \item Compare our solutions with Replika (one of the most advanced commercial social chatbots) as well as with baselines consisting of a system choosing replies randomly, and with a human pretending to be a chatbot.
\end{itemize}

Among the most famous chatbots mentioned in Section \ref{sec:artificial-conv-agents} we chose to use Replika for our experiment as it is available as Android\footnote{Replika's Android application has been downloaded more than 5,000,000 times from the Google Play Store.} and IOS application, and in a web version. Moreover, as already mentioned, it is backed by Open AI's sophisticated GPT-2\footnote{\url{https://en.wikipedia.org/wiki/GPT-2}}: a transformer machine learning model that uses deep learning to generate text output that has been claimed to be indistinguishable from that of humans \cite{cnn-replika}.

\subsection{Participants and Test Groups}
The participants for this test have been recruited through posts on Social Networks (Facebook and Twitter) as well as in Robotics and Computer Science classes at the University of Genoa, for a total of 100 volunteer participants aged between 25 and 65. The only inclusion criteria are that participants shall be able to read and write in English and to use a PC with a network connection from home. Since the test is anonymous and no personal data is collected, the whole procedure is fully compliant with the EU General Data Protection Regulation. Each time a new participant is recruited, he/she is randomly assigned to a group from 1 to 5 (where each group corresponds to a different conversation system), and given the instructions to perform the test. Based on the group they were assigned, participants had to interact with:
\begin{enumerate}
    \item The system exploiting the keyword-based Dialogue Management algorithm;
    \item The system exploiting the keyword and category-based Dialogue Management algorithm;
    \item A system that chooses the next topic randomly (i.e., regardless of what the user says);
    \item A human pretending to be a chatbot (i.e., in a Turing-test fashion);
    \item Replika. 
\end{enumerate}

It shall be reminded that the purpose of the experiments is to test the impact of the Dialogue Management solutions we proposed, which emphasize the problem of switching in-between different topics: Replika is expected to perform better than systems 1, 2, and 3 in generating individual sentences, and this is true in the case of a human pretending to be a chatbot. Then, the purpose of the comparison is to show that, even in presence of a more advanced approach for producing individual sentences, still our solution for Dialogue Management may have a significantly positive impact on user evaluation.

For reasons due to the Covid-19 pandemic, participants could not take the tests with a robot, but they had to interact with their assigned conversational system (1, 2, 3, 4, or 5) using their computer, by establishing a connection to a remote server in our laboratory. 
It is crucial to mention that the text-based user interface was identical for all the systems: therefore, participants assigned group 4 were not aware that they were interacting with a human, writing at a terminal, and participants assigned to group 5 were not aware that they were interacting with Replika (with a human manually copying and pasting sentences from Replika to the terminal and vice-versa).  

In total, each session included 20 exchanges (i.e., a pair of utterances) between the participant and the system.

\subsection{Questionnaire}
Immediately after the conversation session, participants were asked to fill in a questionnaire, divided into two parts. 

The first section of the questionnaire required to evaluate what we defined as the Coherence of system's replies (Section \ref{coherence} below), for a total of 20 replies, individually scored, in a 7-point Likert scale, where 1 means \textit{not Coherent} and 7 means \textit{Coherent}. The second part consisted of the SASSI questionnaire (Section \ref{sassi}).

\subsubsection{Coherence Measure}
\label{coherence}
The instructions provided to each participant clarified what is meant by ``evaluating the Coherence". The aim is to have a way to determine whether the replies of the chatbot are \textit{semantically consistent} with what the user says and/or with what the chatbot itself has previously said. Hence, if the user believes that the reply of the chatbot is perfectly consistent with what has just been said, then the user is instructed to assign 7 to that reply. Otherwise, if the reply of the chatbot has nothing to share with what has just been said in the previous exchange, such a reply should be evaluated with 1. Values in between shall be assigned to replies that are loosely consistent with the current conversation topic or the new topic raised by the user.

The average Coherence score assigned by a participant is computed over all replies; the average Coherence score of a system (i.e., 1, 2, 3, 4, or 5) is computed over all participants that interacted with it.

\subsubsection{SASSI Questionnaire}
\label{sassi}
We decided to use this widely known tool to measure the user experience during the conversation, after performing some research and comparing it with other tools commonly used for this purpose \cite{questionnaires-comparison}.

\begin{figure}
    \centering
    \includegraphics[width=0.9\linewidth]{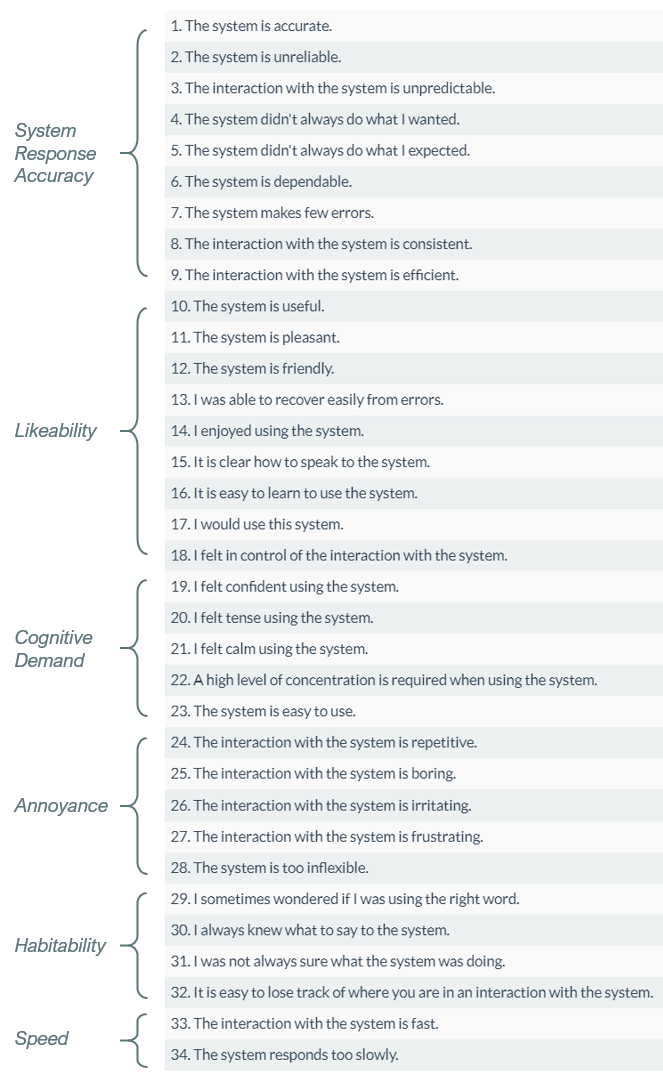}
    \caption{The Subjective Assessment of Speech System Interfaces (SASSI) questionnaire.}
    \label{fig:SASSI}
\end{figure}

Figure \ref{fig:SASSI} reports the Subjective Assessment of Speech System Interfaces (SASSI) questionnaire.

The published version has 34 items distributed across six scales, each item being scored with a 7-point Likert scale from Strongly Disagree to Strongly Agree:
\begin{enumerate}
    \item \textit{System Response Accuracy} (9 items): it refers to the user’s perceptions of the system as accurate and therefore doing what they expect. This will relate to the system’s ability to correctly recognise the speech input, correctly interpret the meaning of the utterance, and then act appropriately;
    \item \textit{Likeability} (9 items): it refers to the user’s ratings of the system as useful, pleasant, and friendly;
    \item \textit{Cognitive Demand} (5 items): it refers to the perceived amount of effort needed to interact with the system and the feelings resulting from this effort;
    \item \textit{Annoyance} (5 items): it refers to the extent to which users rate the system as repetitive, boring, irritating, and frustrating;
    \item \textit{Habitability} (4 items): it refers to the extent to which the user knows what to do and knows what the system is doing;
    \item \textit{Speed} (2 items): it refers to how quickly the system responds to user inputs.
\end{enumerate}

Controversial opinions about the usage of the SASSI can be found in \cite{sassi}.
It has been pointed out that, before being considered as a standard for measuring the quality of speech system interfaces, the SASSI questionnaire should be piloted and validated at a broader scale, and thoroughly revised in the process \cite{Weiss2008} to more firmly establish its psychometric properties. However, to the best of our knowledge, no other validated tools exist to capture the constructs that we want to measure: then, to strengthen the results provided by the SASSI, we introduced the aforementioned Coherence measure, whose correlation with the SASSI scales will be presented later in this Section, by computing Pearson's correlation index based on the collected data.

\subsection{Data Collection}
In total, $100\times 20=2000$ exchanges between the five systems and the participants ($400$ per group), individually scored for their Coherence, have been collected, as well as $100$ completed SASSI questionnaires ($20$ per group). 
To analyze the results of this experiment, an Excel file containing three sheets has been created. 
\begin{itemize}
    \item The first sheet contains all the answers to the first part of the questionnaire, concerning the Coherence of the replies, divided per group (i.e., 1, 2, 3, 4, and 5);
    \item The second sheet contains all the answers to the second part of the questionnaire, divided per group (i.e., 1, 2, 3, 4, and 5) and per SASSI scale. For the Accuracy, Habitability and Speed scales, the scores assigned to negative statements (2, 3, 4, 5, 29, 31, 32, and 34) have been inverted; while for the Cognitive Demand, as a higher score has a negative meaning, we inverted the scores assigned to positive statements (19, 21, 23);
    \item The third sheet contains a statistical analysis of all data. All datasets have been tested for normality both with the Shapiro-Wilk Test and with criteria based on the descriptive statistics (data are normally distributed if the absolute values of both Skewness and Kurtosis are $\le 1$).
\end{itemize}

The internal consistency of SASSI data has been checked by computing the Cronbach's alpha for all scales: for Accuracy, alpha=0.91; for Likeability, alpha=0.91; for Cognitive Demand, alpha=0.60, for Annoyance, alpha=0.81; for Habitability, alpha=0.64; for Speed, alpha=0.92. Most scales/groups present a reliability of more than 0.80, ranging from good to excellent. However, Cognitive Demand and Habitability present a reliability in the range of 0.6-0.7, which is considered questionable.
We hypothesize that low alpha values can be due to the fact that we did not recruit only native English speakers for our experiments, since our inclusion criteria only required that participants could read and write in English: the fact that Cognitive Demand and Habitability have some negative and/or complex questions may have had an impact on the internal consistency of the collected scores. Results related to such scales are reported in Section \ref{sec:results}, even if they should be taken \textit{cum grano salis}.

The average scores of every group/scale are computed and pairwise compared. To assess whether there is a significant difference between the datasets, the \textit{Mann-Whitney U test} has been used: this test is an alternative to the t-test when data are not normally distributed. In case the comparison is performed among two normally distributed samples, the \textit{Welch's t-test} has been additionally performed.

The Ontology used by systems 1, 2, and 3, including all concepts as well as related sentences and keywords, the resulting DT with topics of conversation, the mapping between CNL categories and topics in the DT, and finally Excel files with individual replies to questionnaires as well as data analysis are openly available\footnote{ \url{http://caressesrobot.org/IJSORO2021/}}

\section{Results}
\label{sec:results}
This section presents the results obtained for the considered seven items: the Coherence and the six scales of the SASSI.
First, we computed the mean value and the standard deviation for each group, with the aim of comparing them two by two, making the null hypothesis that the compared groups are equal. To verify whether such hypothesis shall be rejected, we performed the Shapiro-Wilk test for normality on each of the five datasets: if the distribution was not normal, we performed only the Mann-Whitney U test, while if the distribution was normal, we performed both the Mann-Whitney U test and the Welch's t-test. In case both tests were performed, we compared the p-values to verify if the outcomes were consistent. 

For each item, we report a histogram with the mean values and the standard deviations, and a table with the U-value and U-critical (corresponding to $p=0.05$ for the Mann-Whitney U test), and the p-values (computed with both tests, when appropriate). The green cells highlight the cases when the null hypothesis is rejected, with $p < 0.05$, i.e., there is a significant difference between the groups (in many cases, we found $p < 0.01$). 

\begin{figure}
    \centering
    \includegraphics[width=\linewidth]{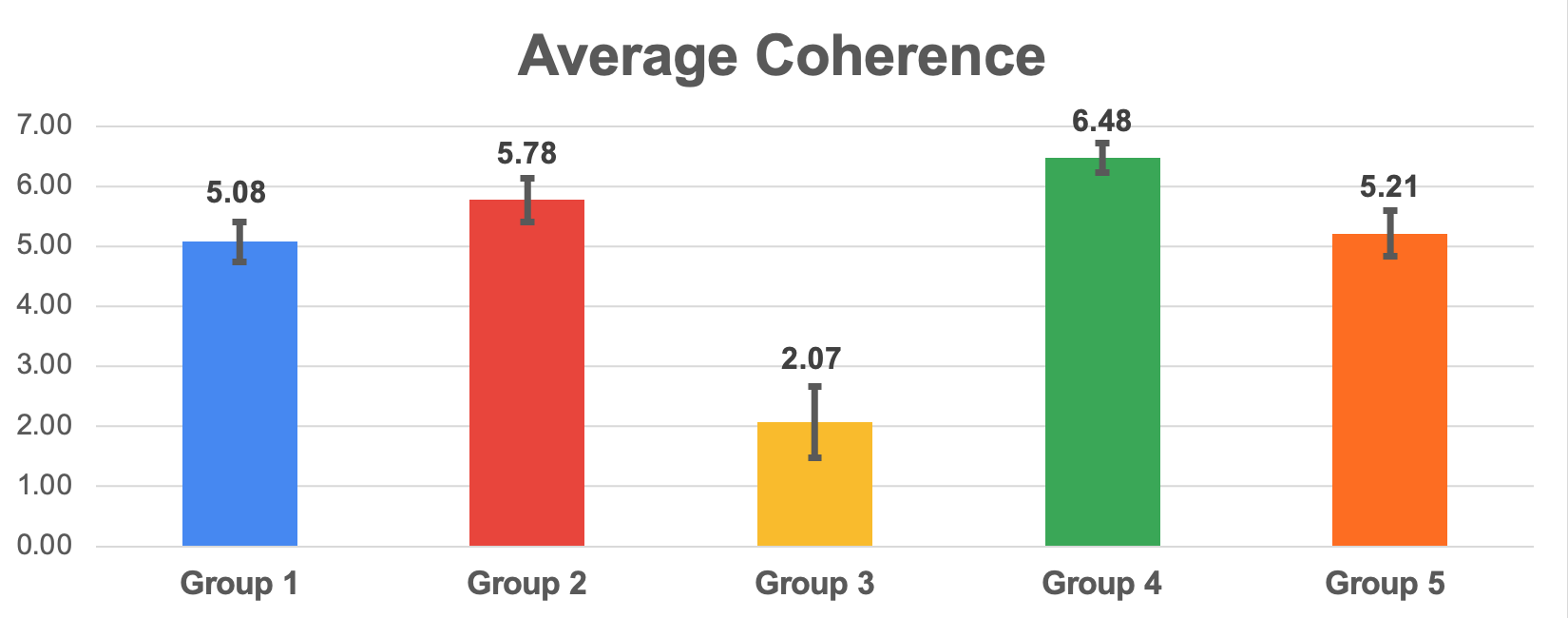}
    \caption{Histogram reporting the average Coherence and the standard deviation of each group.}
    \label{fig:average-coherence}
\end{figure}

\begin{figure}
    \centering
    \includegraphics[width=\linewidth]{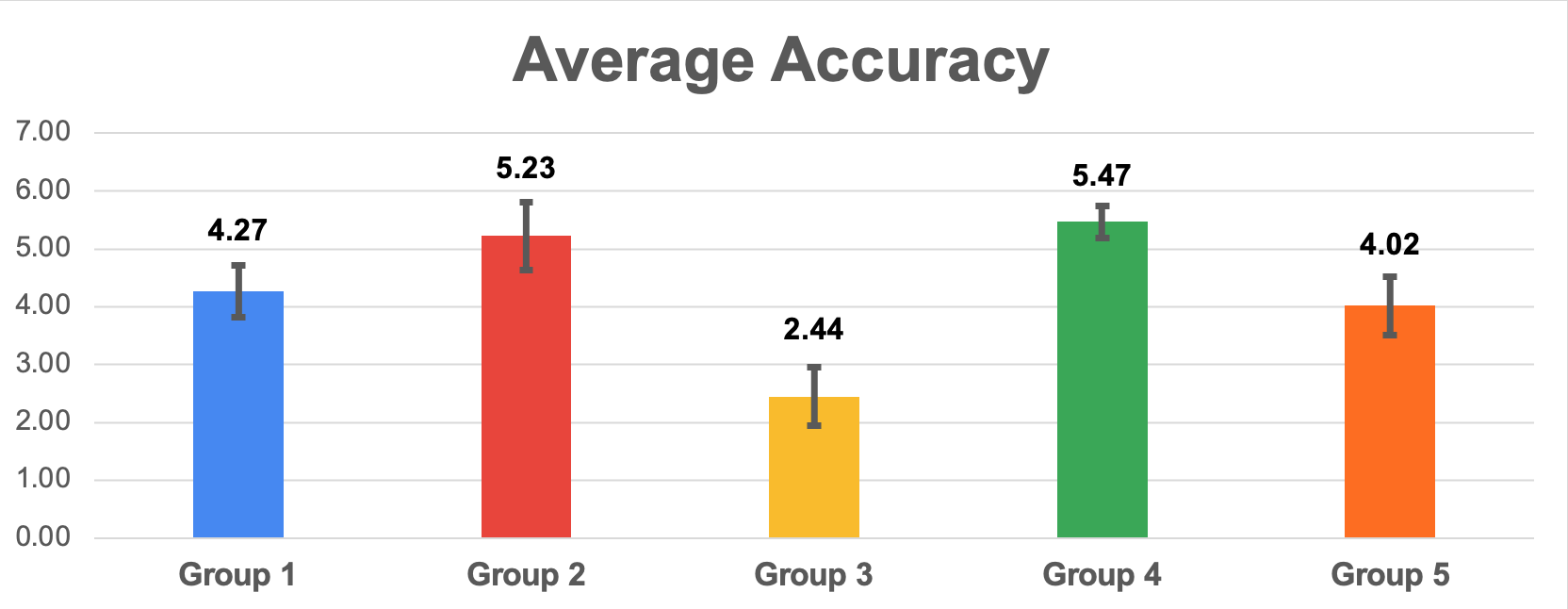}
    \caption{Histogram reporting the average Accuracy and the standard deviation of each group.}
    \label{fig:average-accuracy}
\end{figure}

\begin{figure}
    \centering
    \includegraphics[width=\linewidth]{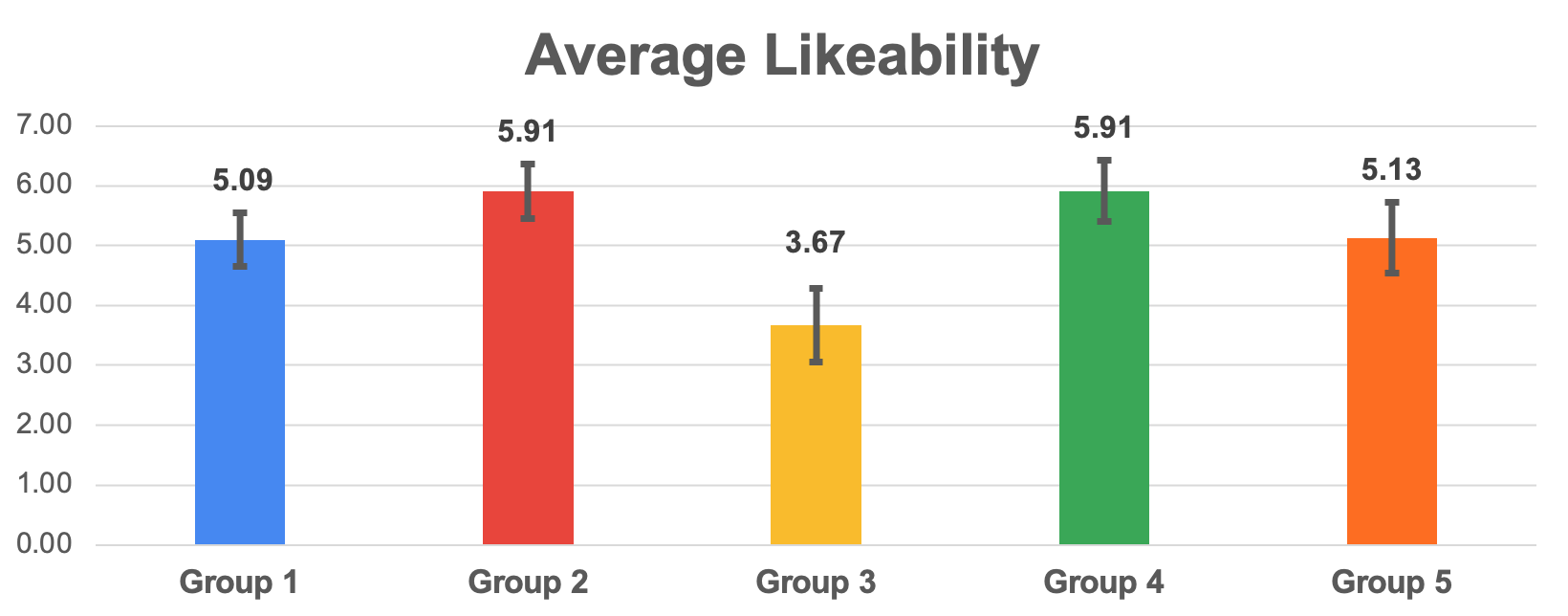}
    \caption{Histogram reporting the average Likeability and the standard deviation of each group.}
    \label{fig:average-likeability}
\end{figure}

\begin{figure}
    \centering
    \includegraphics[width=0.92\linewidth]{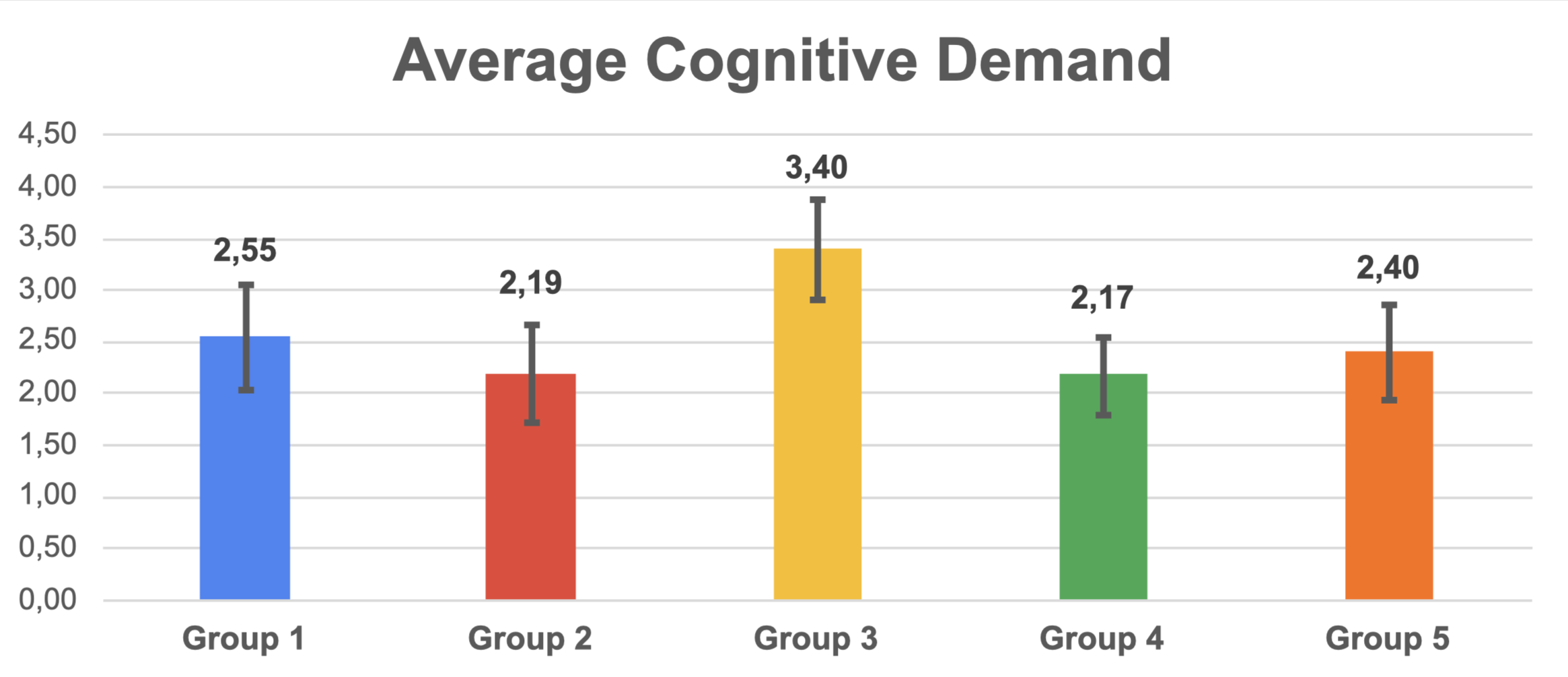}
    \caption{Histogram reporting the average Cognitive Demand and the standard deviation of each group.}
    \label{fig:average-cognitive}
\end{figure}

\begin{figure}
    \centering
    \includegraphics[width=0.99\linewidth]{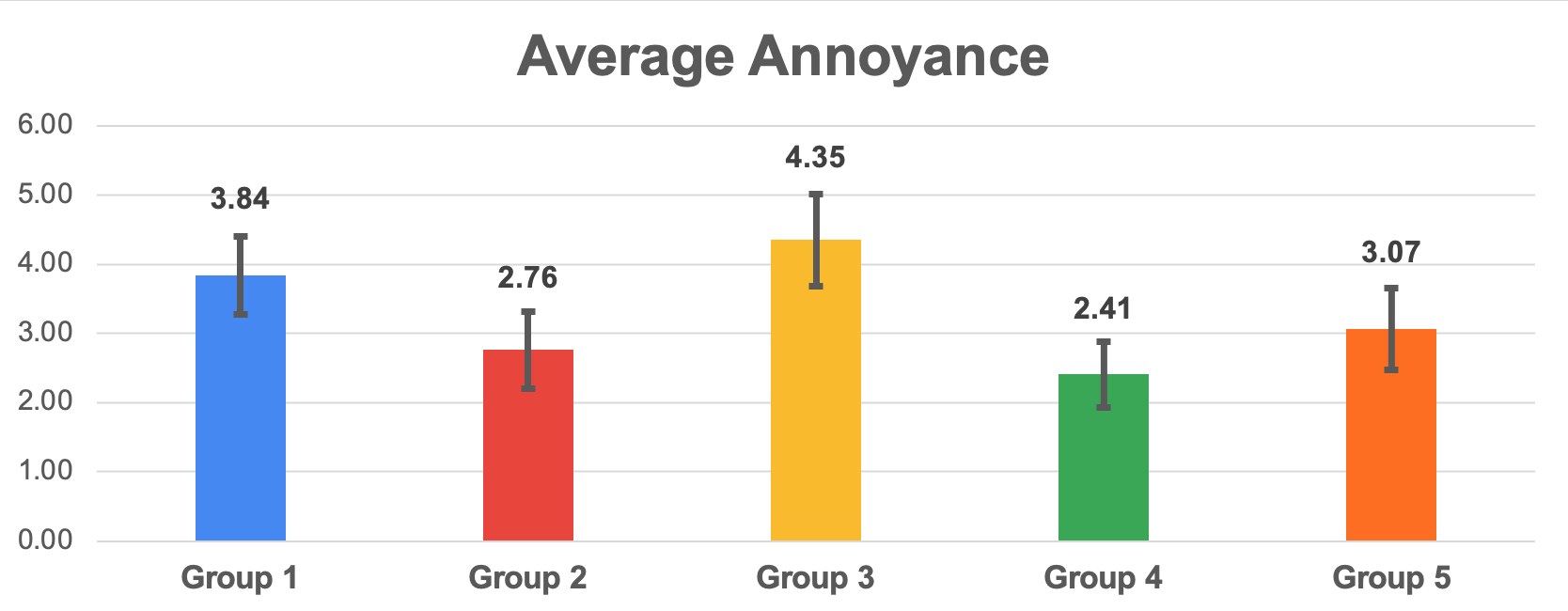}
    \caption{Histogram reporting the average Annoyance and the standard deviation of each group.}
    \label{fig:average-annoyance}
\end{figure}

\begin{figure}
    \centering
    \includegraphics[width=\linewidth]{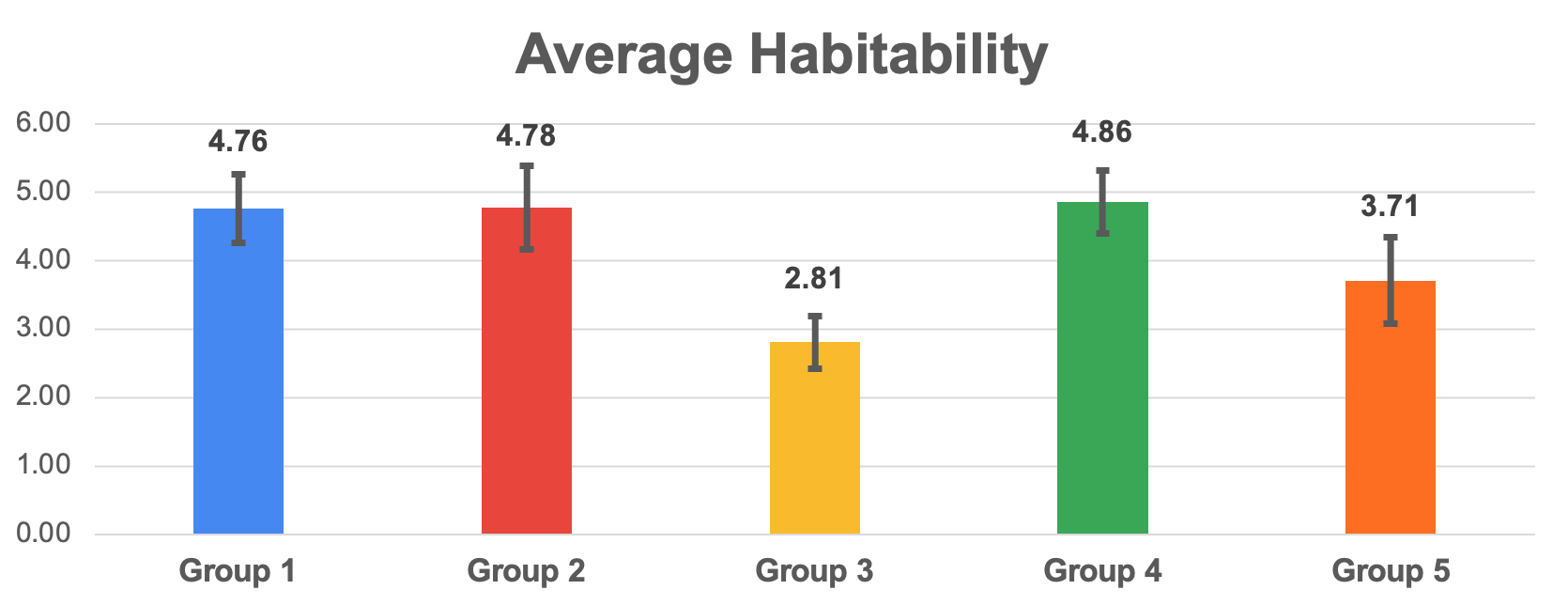}
    \caption{Histogram reporting the average Habitability and the standard deviation of each group.}
    \label{fig:average-habitability}
\end{figure}

\begin{figure}
    \centering
    \includegraphics[width=\linewidth]{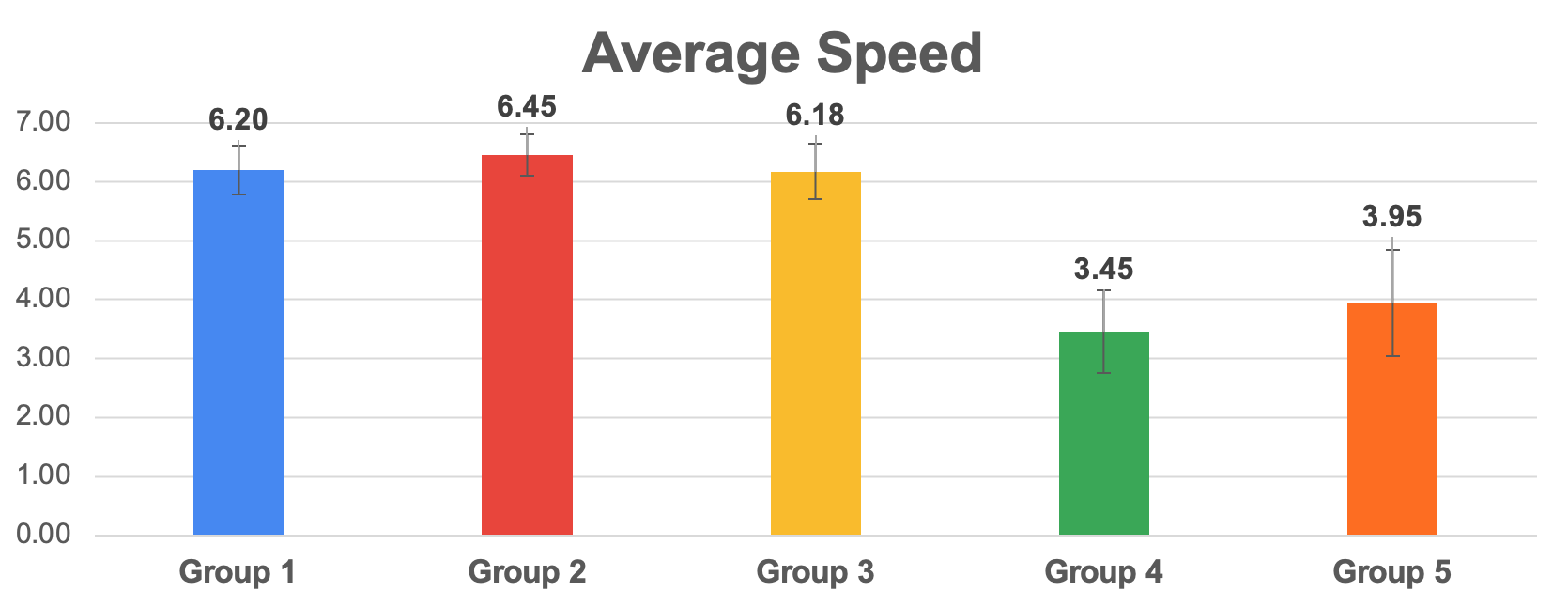}
    \caption{Histogram reporting the average Speed and the standard deviation of each group.}
    \label{fig:average-speed}
\end{figure}

Figures \ref{fig:average-coherence}, \ref{fig:average-accuracy}, \ref{fig:average-likeability}, \ref{fig:average-cognitive}, \ref{fig:average-annoyance}, \ref{fig:average-habitability}, and \ref{fig:average-speed}, report the histograms representing the average and the standard deviation for each group respectively for the seven items, i.e., Coherence, Accuracy, Likeability, Cognitive Demand, Annoyance and Speed. Please notice that, in contrast with the other scales of the SASSI, a higher value of Cognitive Demand represents a negative aspect. 

\begin{figure}
    \centering
    \includegraphics[width=\linewidth]{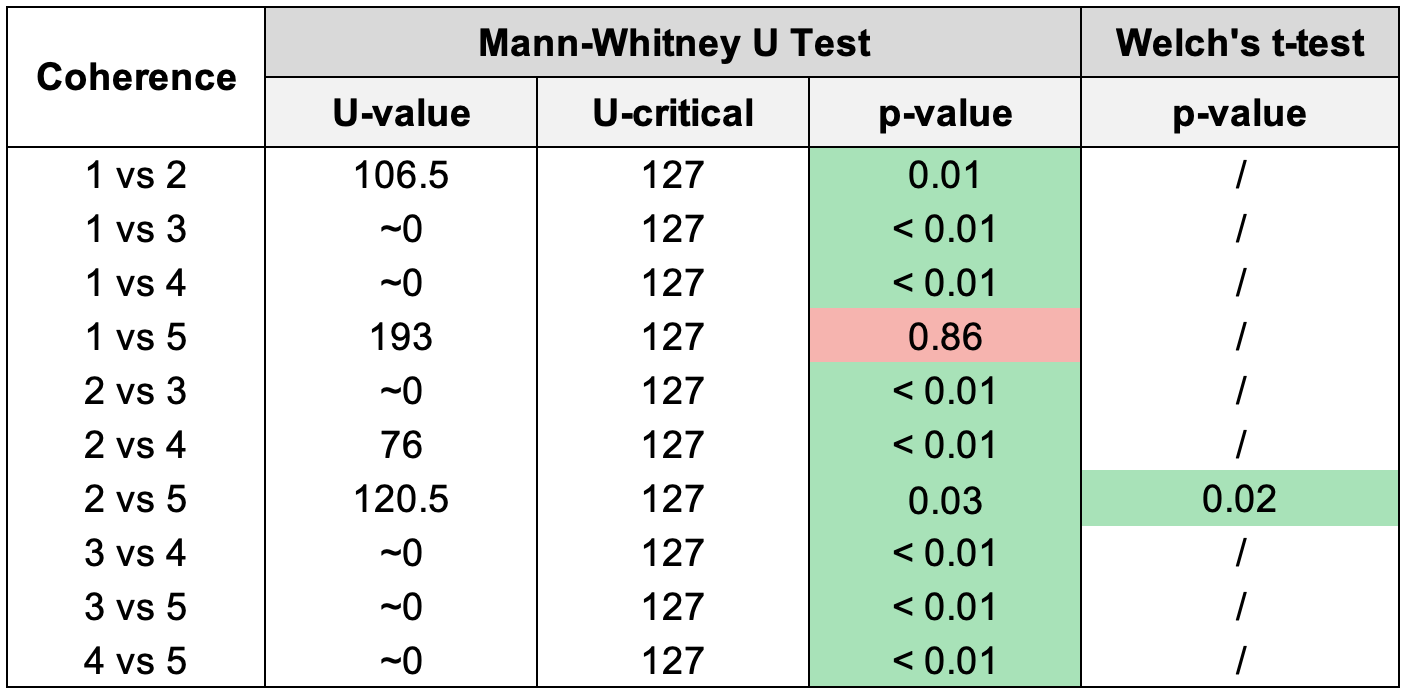}
    \caption{Pairwise statistical comparison of Coherence for different groups.}
    \label{fig:p-value-coherence}
\end{figure}

\begin{figure}
    \centering
    \includegraphics[width=\linewidth]{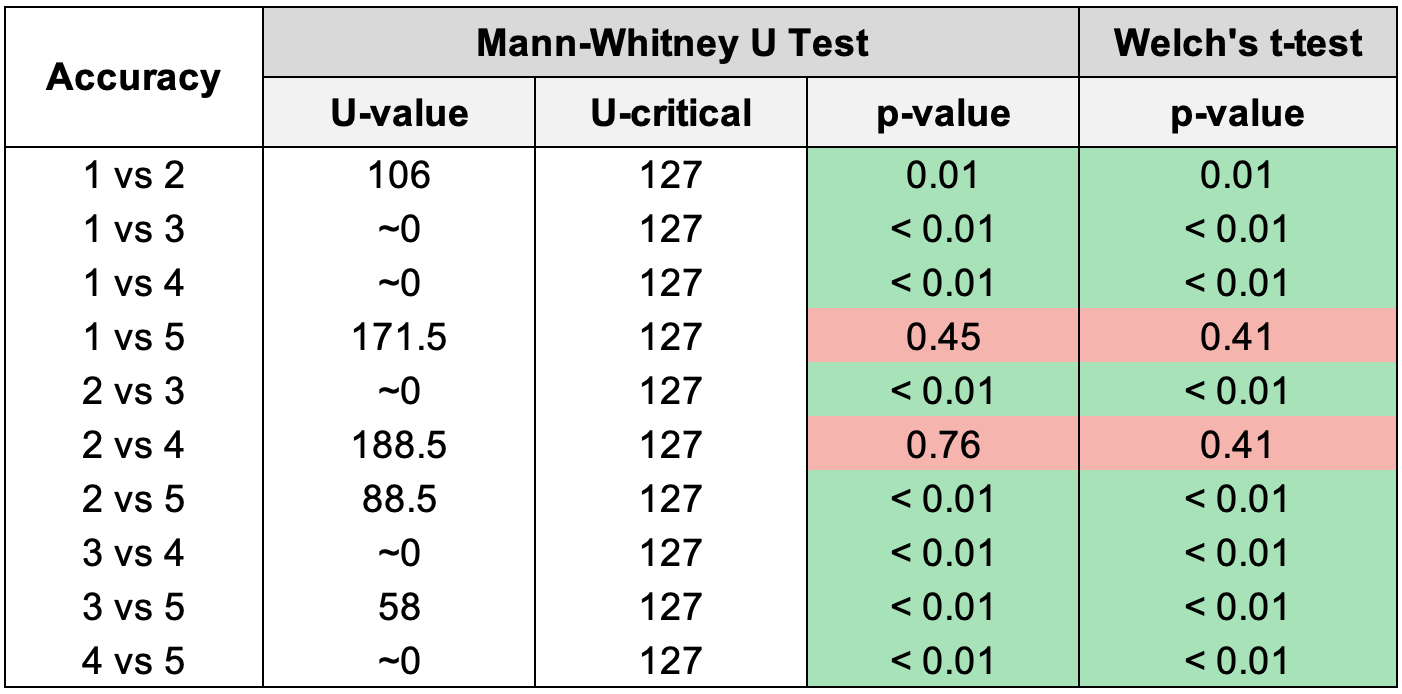}
    \caption{Pairwise statistical comparison of Accuracy for different groups.}
    \label{fig:p-value-accuracy}
\end{figure}

\begin{figure}
    \centering
    \includegraphics[width=\linewidth]{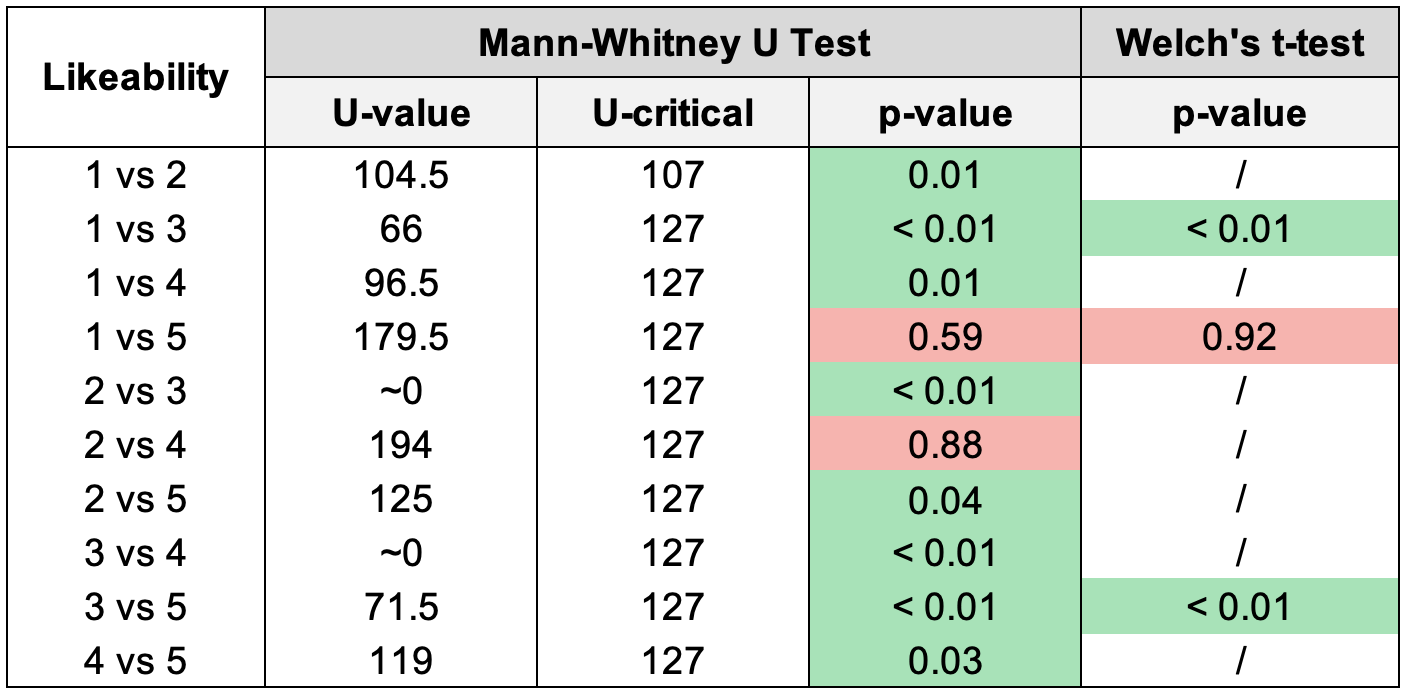}
    \caption{Pairwise statistical comparison of Likeability for different groups.}
    \label{fig:p-value-likeability}
\end{figure}

\begin{figure}
    \centering
    \includegraphics[width=\linewidth]{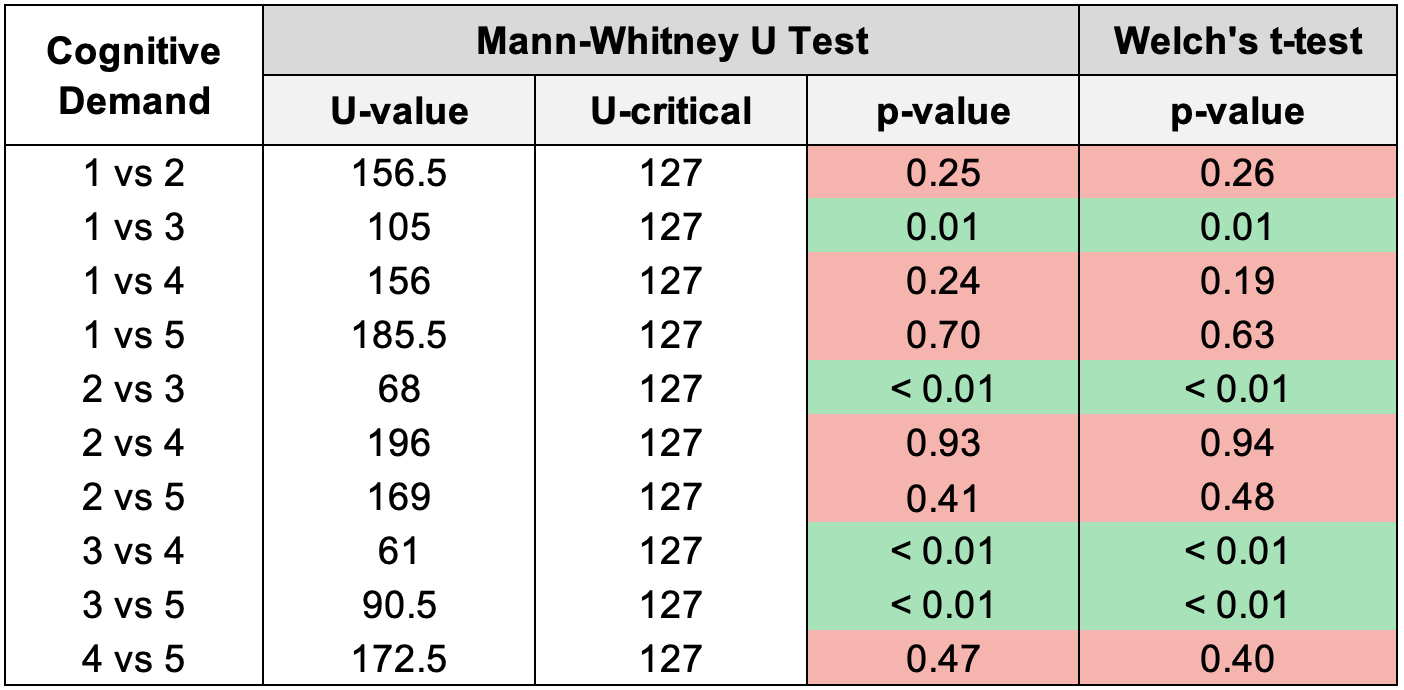}
    \caption{Pairwise statistical comparison of Cognitive Demand for different groups.}
    \label{fig:p-value-cognitive}
\end{figure}

\begin{figure}
    \centering
    \includegraphics[width=0.99\linewidth]{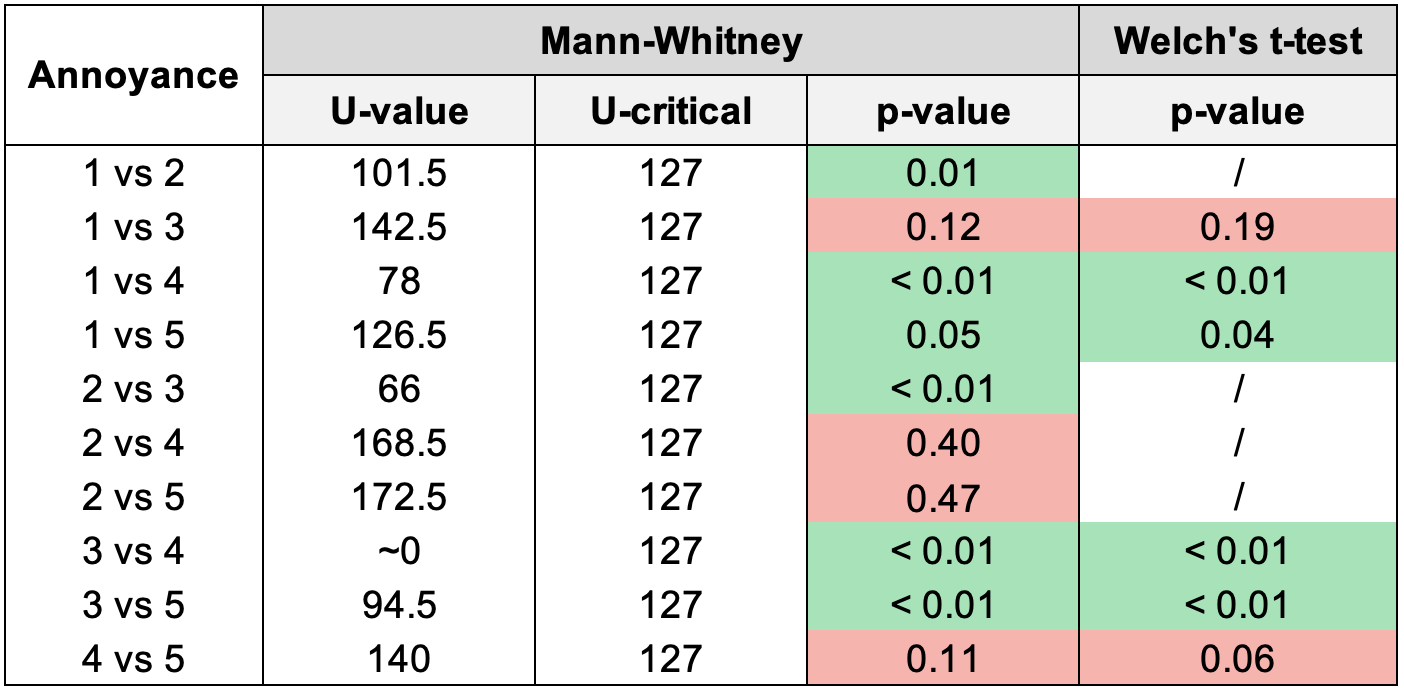}
    \caption{Pairwise statistical comparison of Annoyance for different groups.}
    \label{fig:p-value-annoyance}
\end{figure}

\begin{figure}
    \centering
    \includegraphics[width=\linewidth]{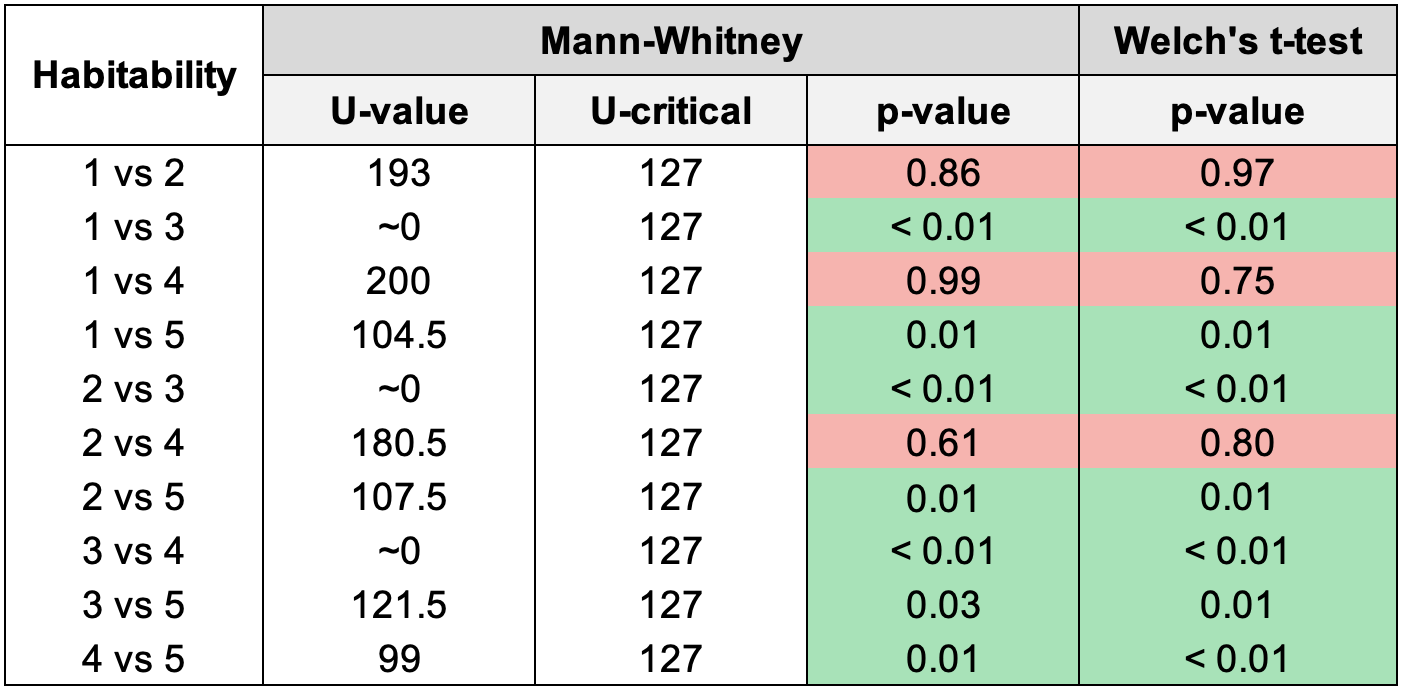}
    \caption{Pairwise statistical comparison of Habitability for different groups.}
    \label{fig:p-value-habitability}
\end{figure}

\begin{figure}
    \centering
    \includegraphics[width=\linewidth]{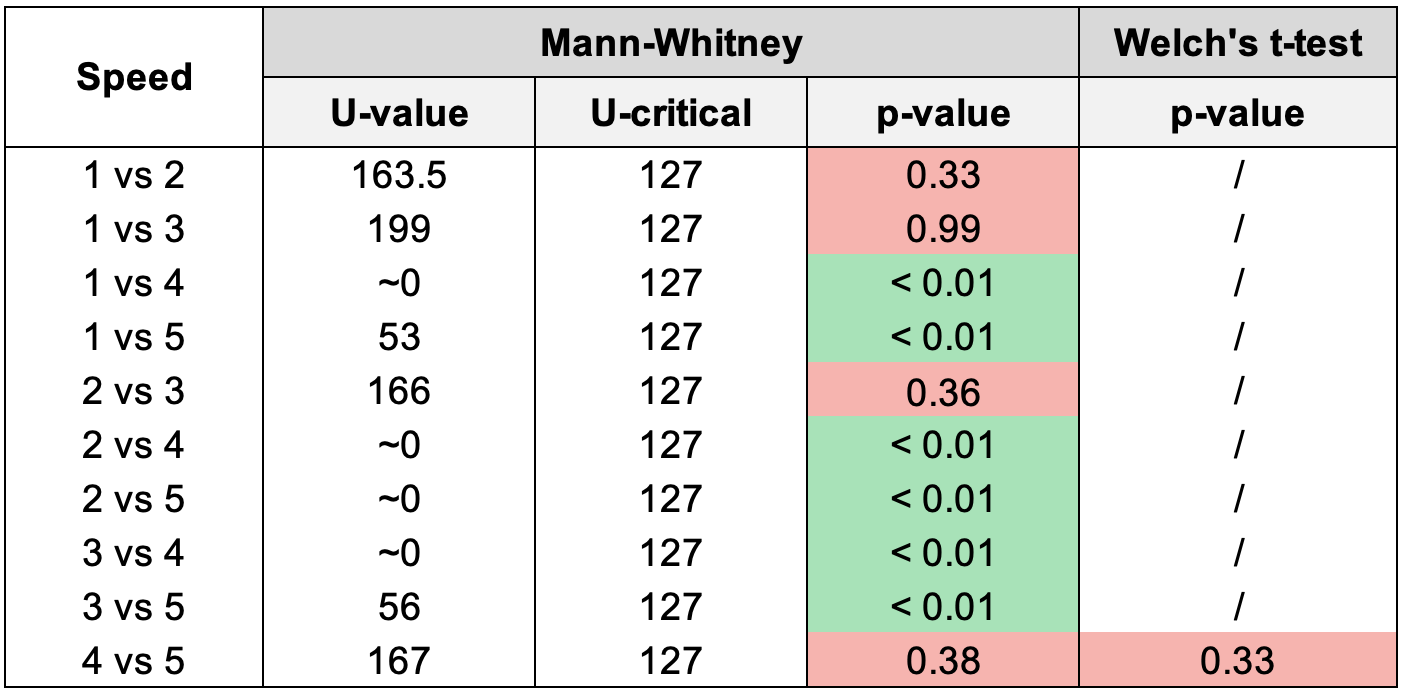}
    \caption{Pairwise statistical comparison of Speed for different groups.}
    \label{fig:p-value-speed}
\end{figure}

Figures \ref{fig:p-value-coherence}, \ref{fig:p-value-accuracy}, \ref{fig:p-value-likeability}, \ref{fig:p-value-cognitive}, \ref{fig:p-value-annoyance}, \ref{fig:p-value-habitability}, and \ref{fig:p-value-speed}, present the tables containing the values computed through the Mann-Whitney U test and Welch’s t-test. Note that, when both statistical tests are performed, the p-values are always consistent. 

\subsection{Correlation between Coherence and SASSI scales}
The correlation between Coherence measures and each SASSI scale has been computed by considering the average score totalled by each participant (yielding 100 values for Coherence and 100 for each SASSI scale), and computing the Pearson's correlation index by pairing Coherence measures with the corresponding scores of each SASSI scale. The value of the correlation found when comparing Coherence with Accuracy turns out to be r=0.78; with Likeability r=0.69; with Cognitive Demand r=-0.45; with Annoyance r=-0.51; with Habitability r=0.56; with Speed r=-0.24. As expected, a high positive correlation is reported between Coherence and Accuracy and, to a minor extent, with Likeability and Habitability. A weaker, negative correlation is found with Cognitive Demand and Annoyance. An easily explainable small negative correlation with Speed is reported: system 4 (i.e., a human pretending to be a chatbot) obtains the highest Coherence scores but needs a longer time to reply, whereas system 3 (i.e., the one that chooses topics randomly) obtains the lowest Coherence scores but has no delays when answering since no reasoning algorithms are involved.

\section{Discussion}
\label{sec:discussion}
This section highlights and examines the major findings.

\subsection{Coherence}
Figure \ref{fig:average-coherence} shows that the standard deviations of groups 1, 2, and 5 are very similar and not too big: this means that, in general, the evaluations regarding the Coherence were quite homogeneous. Looking at the standard deviation of group 3, it is clear that the scores assigned by participants to its Coherence were more dissimilar. This may be because the random answers provided by the system during this test are more coherent during some conversations rather than others, or because some participants may be biased to interpret sentences to re-establish coherence with the context. Eventually, the standard deviation of group 4 is the smallest one: participants, without knowing that they were interacting with a real human, gave a homogeneously positive evaluation of the Coherence of the system's replies.

From Figure \ref{fig:p-value-coherence}, and by looking at the averages of the Coherence reported in the histogram in Figure \ref{fig:average-coherence}, we can state that, as expected, group 3 has the lowest Coherence (with a significant difference with all the others), and group 4 has the highest Coherence (again with a significant difference from the others). The statistical analysis revealed a significant difference between groups 1 and 2 and between groups 2 and 5.

Results confirm that the additional cost for category extraction and matching (2) positively impacts the perceived Coherence with respect to using keywords only (1), and it is sufficient for beating Replika (5).   

\subsection{Accuracy}
Figure \ref{fig:average-accuracy} shows that the standard deviations of the Accuracy corresponding to different groups are very similar, except for the one of group 4 which is lower than the others: this indicates that the participants evaluated more homogeneously the Accuracy of this ``system" (the human).

As for Coherence, the random system (3) is the worst one. An interesting result is that the system exploiting the keyword and category-based Dialogue Management algorithm (2) is the only one as accurate as a human (4) (no statistically significant difference found), other than being remarkably more accurate than all the other systems. 
\subsection{Likeability}
The standard deviations shown in Figure \ref{fig:average-likeability} are very similar; group 3 reports a slightly higher standard deviation, which indicates that participants had different opinions regarding the utility and the pleasantness of the systems. As happened for the Coherence (Section \ref{coherence}), the higher standard deviation corresponding to group 3 may be related to chance, as well as to biases of participants to favourably interpret random replies to re-establish missing coherence.

As it could be expected, from the histograms it is immediate to notice that group 3 has the lowest Likeability. As regards the other groups, the average Likeability scores present less prominent differences.

Considering both the table in Figure \ref{fig:p-value-likeability} and the histogram in Figure \ref{fig:average-likeability}, the most interesting result is that the system that exploits the keyword and category-based Dialogue Management algorithm (2), is significantly more likeable than the one exploiting keywords only (1) and Replika (5). Moreover, the Likeability of (2) is similar to that of a human (4), while this is not true when comparing the human with the system using keywords only (1), fully justifying the additional cost for category extraction and matching. 

\subsection{Cognitive Demand}
The standard deviations in Figure \ref{fig:average-cognitive} are again very similar; group 4 reports a slightly lower standard deviation which, as always, indicates that the participants' evaluation regarding the Cognitive Demand of the system is homogeneous.

As regards the averages, group 3 deviates from the others. This means that a higher cognitive effort is needed to interact with the system: this is since the replies of the system are random, hence the participants found it more difficult to easily interact with it.

By looking at the p-values in Figure \ref{fig:p-value-cognitive} and the averages in Figure \ref{fig:average-cognitive}, we can conclude that all the systems require the same amount of effort during the interaction, except the one providing random replies (3). A valuable result is that both systems we developed (1 and 2) present no significant difference, in terms of Cognitive Demand, with respect to a human (4) and Replika (5). 

\subsection{Annoyance}
As shown in Figure \ref{fig:average-annoyance}, the standard deviations are quite high and similar to one another. Again, the lowest standard deviation corresponds to group 4: when participants unknowingly interacted with a human, their ratings were more homogeneous. Considering the averages, as we could expect, the system used with group 3 appears to be the most annoying, while the lowest Annoyance score is associated with group 4. 

Examining the table in Figure \ref{fig:p-value-annoyance} and the histogram in Figure \ref{fig:average-annoyance}, we can conclude that system 2 (keywords plus categories) is significantly less annoying than 1 (keywords only), and it is as annoying as 4 (human) and 5 (Replika). Notice also that 1 (keywords only) is significantly more annoying than both 4 (human) and 5 (Replika).

\subsection{Habitability}
From the histograms in Figure \ref{fig:average-habitability} it can be seen that the standard deviations regarding the Habitability are very similar for all groups, with the smallest one being that of group 3.

Regarding the averages, the smallest one is that corresponding to group 3, which presents also the lowest average Habitability. This result was expected, as the random system had also the lowest Accuracy, Likeability, and Annoyance, and the highest Cognitive Demand, whose scores are likely not completely independent of one another.

Considering the p-values in Figure \ref{fig:p-value-habitability}, and the averages in Figure \ref{fig:average-habitability}, it can be observed that the system used with group 2 (keywords plus categories) turned out to be more habitable than 5 (Replika), while 1 (keywords only) is less habitable than 5. Very interestingly, both our systems 1 and 2 are not distinguishable from 4 (human) in Habitability.

\subsection{Speed}
As it can be seen in Figure \ref{fig:average-speed}, the system tested with group 3 (random replies) is faster than 1 and 2, since they are all connected to the CARESSES Cloud, but the former does not call the Dialogue Management algorithm.
However, the standard deviation of group 3 is slightly higher than those of groups 1 and 2, even if the average is lower: we conjecture that the extremely low pleasantness of the overall interaction with this system, confirmed by all the results of the previous sections, negatively influences the perception of Speed.

Groups 4 and 5 report a higher standard deviation and a lower average. However, these results cannot be compared with those of the other tests. Since group 4 involves the unaware interaction with a human, the final score depends on the typing speed of the human.
The same reasoning applies to group 5, which involved the unaware interaction with Replika: a human acted as an intermediary between the participant and Replika, by manually typing Replika's replies which required some additional time.

Considering both the table in Figure \ref{fig:p-value-speed}, and the histogram in Figure \ref{fig:average-speed}, the only relevant result is that there are no significant Speed differences between 1, 2, and 3. No conclusions shall be drawn when comparing groups 4 and 5.

\section{Conclusion}
\label{sec:conclusion}
The article proposes a novel system for knowledge-based conversation designed for Social Robots and other conversational agents. The proposed system relies on an Ontology for the description of all concepts that may be relevant conversation topics, as well as their mutual relationships. We compare two algorithms, based on the Ontology, for Dialogue Management that select the most appropriate conversation topics depending on the user's input: the two versions differ in their computational cost and/or the need for third-party NLP services. Moreover, they implement slightly different strategies to ensure a conversation flow that captures the user's intention to drive the conversation in specific directions, while avoiding purely reactive responses.

Experiments performed with 100 volunteer participants, interacting with five different conversational systems (one of which is a human pretending to be a chatbot in a Turing-test fashion), support our intuitions about the importance of dialogue flow management, the improvements brought by the proposed solution based on the semantic category of the user's sentence, as well as the positive correlations between Coherence in flow management and some scales of the SASSI (Subjective Assessment of Speech System Interfaces) questionnaire.
Specifically, one of the proposed algorithms is statistically superior (and never inferior), in all items, to Replika: one of the most popular chatbots worldwide.

The approach presented in this article has obvious limitations since it relies on an Ontology of concepts and the related sentences to talk about such concepts, which needs to be manually encoded by experts. This issue has been addressed in subsequent work, which explored strategies for expanding the knowledge base at run-time during the interaction with the user \cite{Grassi2021}.

\section*{Conflict of Interest}
The authors declare that they have no conflict of interest.

\bibliographystyle{unsrt}      
\bibliography{bibliography} 


\vspace{10mm}
\textbf{Lucrezia Grassi} (M.D.) is a Ph.D. student in Bioengineering and Robotics at the University of Genova. She got her master's degree in Robotics Engineering in 2020 with the thesis ``A Knowledge-Based Conversation System for Robots and Smart Assistants", and she is currently pursuing her Ph.D. on multiparty interaction between humans and artificial agents. Her interests include Social Robotics, autonomous conversation systems, mixed and virtual reality.

\vspace{10mm}
\textbf{Carmine Tommaso Recchiuto} (Ph.D.) is Assistant Professor at the University of Genoa, where he teaches Experimental Robotics, ROS programming, and Computer Science. His research interests include Humanoid and Social Robotics (with a specific focus on knowledge representation and human-robot interaction), wearable sensors, and Aerial Robotics. He has been the Coordinator of software integration and Head of Software Development for the CARESSES project, aimed at endowing social robots for older adults with cultural competence. He has also been the local Coordinator for the BrainHuRo project, developing Brain-Computer Interfaces for humanoid robots' remote control. He is the author of more than 40 scientific papers published in International Journals and conference proceedings.

\vspace{10mm}
\textbf{Antonio Sgorbissa} (Ph.D.) is Associate Professor at the University of Genoa, where he teaches Real-Time Operating Systems, Social Robotics, and Ambient Intelligence in EMARO+, the European Master in Advanced Robotics. He is the Coordinator of National and EU research projects, among which the H2020 project CARESSES (\url{caressesrobot.org}). Also, he is the local Coordinator of the ongoing IENE-10 project, aimed at preparing health and social care workers to work with intelligent robots in health and social care environments. He received a number of acknowledgements for his recent work: among the others, CARESSES has been acknowledged "Project of the month" by the EU, its technologies have been acknowledged by the EU Innovation Radar, and the project has been included among the ``100 Italian Robotics \& Automation Stories” in a report presented by Enel S.p.a. in February 2020.
His research focuses on Autonomous Robotic Behaviour, with a focus on Culture-Awareness, Knowledge representation, Motion Planning, Wearable, and Ubiquitous Robotics.
He is the author of about 150 articles indexed in international databases and has been a member of the Organizing Committee in the top-ranked robotic conference as well as Associate Editor for the International Journal of Advanced Robotic Systems edited by SAGE and Intelligent Service Robotics by Springer.

\end{document}